\documentclass[letterpaper, 10pt, conference]{ieeeconf}
\IEEEoverridecommandlockouts
\usepackage{epsfig}
\usepackage{amssymb}
\usepackage{cite}
\usepackage{booktabs}
\usepackage{multirow}
\usepackage{mathtools}
\usepackage{bm}

\newcommand{\figlab}[1]{\label{fig:#1}}
\newcommand{\figref}[1]{Fig.~\ref{fig:#1}} % Figure
\newcommand{\tablab}[1]{\label{tab:#1}}
\newcommand{\tabref}[1]{Table~\ref{tab:#1}} % Table
\newcommand{\forlab}[1]{\label{for:#1}}
\newcommand{\forref}[1]{Equation~(\ref{for:#1})} % Equation

\newcommand{\etal}{\textit{et~al.}}

\newcommand{\eg}{\textit{e.g.,}}

\usepackage{subcaption}
\usepackage{url}
\usepackage{threeparttable}

\usepackage{amsmath}
\usepackage{algorithmic}
\usepackage{array}
\title{\LARGE \textbf{Assembly Sequences Based on Multiple Criteria \\Against Products with Deformable Parts}}
\author{Takuya Kiyokawa, Jun Takamatsu, and Tsukasa Ogasawara%
\thanks{All authors are with the Division of Information Science, Robotics Laboratory, Nara Institute of Science and Technology (NAIST), Japan {\tt\small \{kiyokawa.takuya.kj5, j-taka, ogasawar\}}@is.naist.jp}}

\begin{document}

\maketitle
\thispagestyle{empty}
\pagestyle{empty}

\begin{abstract}
To generate assembly sequences that robots can easily handle, this study tackled assembly sequence generation (ASG) by considering two tradeoff objectives: (1) insertion conditions and (2) degrees of the constraints affecting the assembled parts. We propose a multiobjective genetic algorithm to balance these two objectives. 
Furthermore, we extend our previously proposed 3D computer-aided design (CAD)-based method for extracting three types of two-part relationship matrices from 3D models that include deformable parts. 
The interference between deformable and other parts can be determined using scaled part shapes. 
Our proposed ASG can produce Pareto-optimal sequences for multi-component models with deformable parts such as rubber bands, rubber belts, and roller chains. 
We further discuss the limitation and applicability of the generated sequences to robotic assembly.
\end{abstract}

\section{Introduction}
Production systems that can respond quickly to changes in market demands are needed~\cite{Gunasekaran2019}. 
For such agile manufacturing~\cite{Gunasekaran1999,Costa2017}, assembly sequences must be generated rapidly. 
Several studies for assembly sequence generation (ASG) use 3D computer-aided design (CAD) models~\cite{Bahubalendruni2015,Lee2016,Deepak2019}.

The combinatorial optimization problem for the ASG~\cite{Jimenez2013} is known to be NP-hard~\cite{Goldwasser1999}. 
To obtain quasi-optimal solutions in realistic time, heuristic search methods have been used. 
Some researchers used genetic algorithms (GAs)~\cite{Smith2001,Chen2001,Smith2002} for the ASG in two dimensions.
Pan~\etal~\cite{Pan2006} generated multiple sequences from only a 3D CAD file; however, the final sequence had to be determined manually.

Tariki~\etal~\cite{Tariki2019} set out to generate preferable sequences for robots by initializing the chromosomes of GA based on the interference between many parts (\eg~32). 
They used insertion relationships (\eg~plug-receptacle, peg-hole, and pin-slot)~\cite{Tariki2020} and defined \textit{preferable insertion sequence condition} (hereinafter referred to as ``insertion condition'').

However, as shown in~\figref{issue}, the insertion sequence generated by the method~\cite{Tariki2020} causes simultaneous contact between multiple parts. 
Such insertions are difficult to handle. 
%%%%%%%%%%%%%%%%%%%%%%%%%%%%%%%%%%
\begin{figure}[tb]
    \centering
    \includegraphics[width=0.65\linewidth]{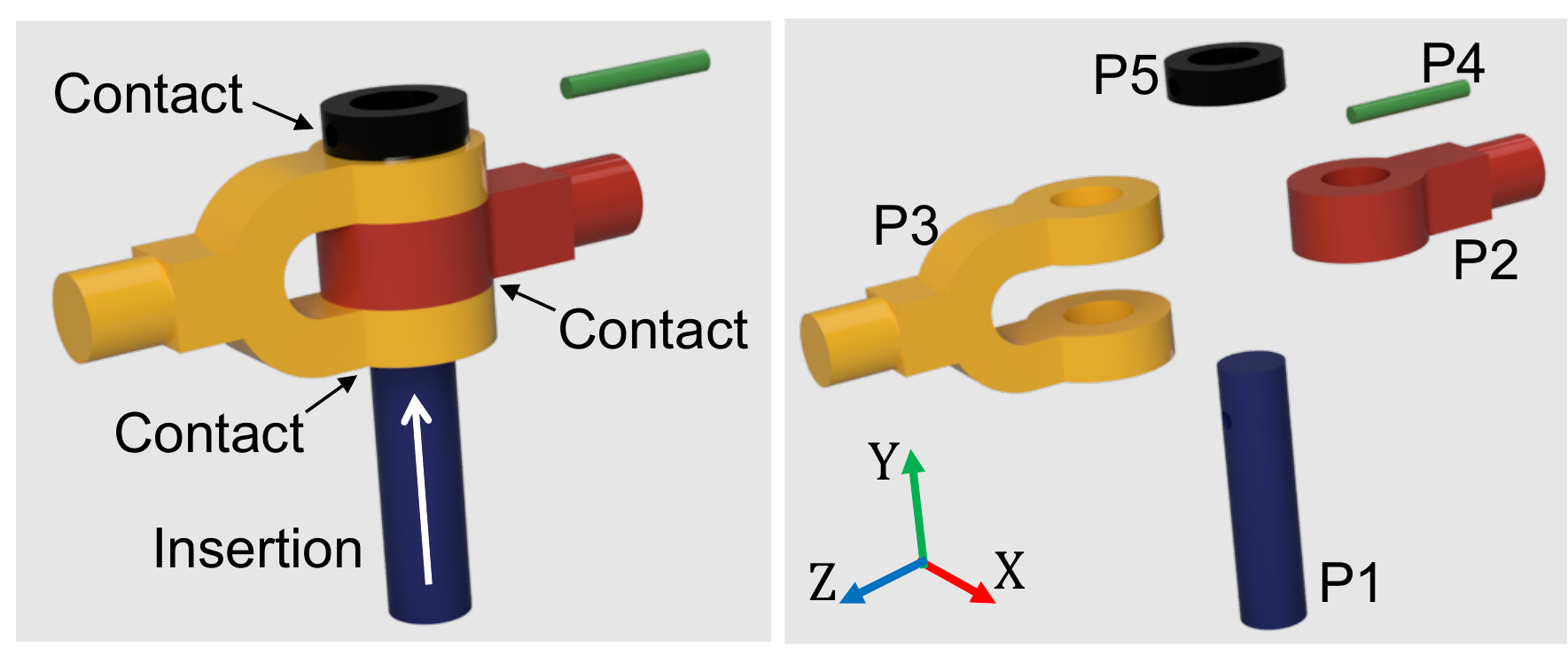}
    \caption{\small{Insertion sequence (white arrows) for part that creates three contact surfaces between the parts of the model (\#1).}}
    \figlab{issue}
\end{figure}
%%%%%%%%%%%%%%%%%%%%%%%%%%%%%%%%%%
\begin{figure*}[tb]
    \centering
    \includegraphics[width=0.8\linewidth]{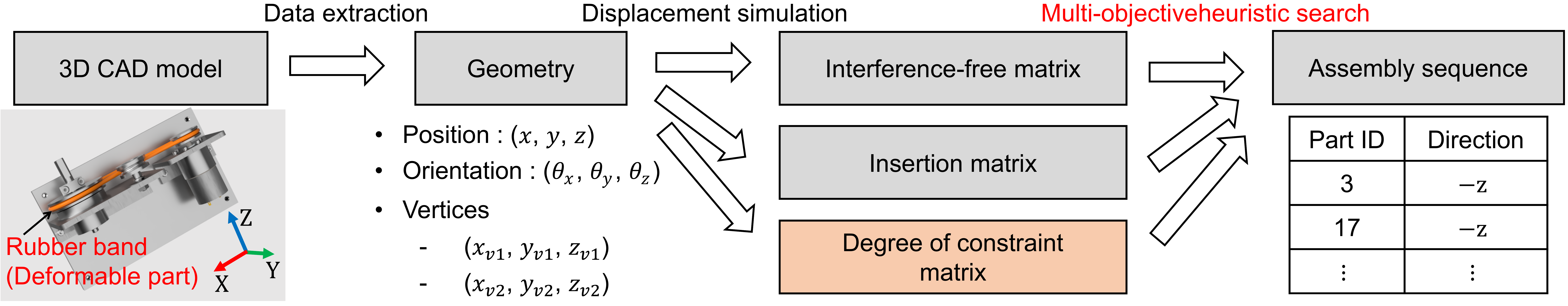}
    \caption{\small{Overview of generation of assembly sequence from a 3D CAD model. The input is an assembled-3D CAD model and the output is a serial assembly sequence of part IDs and the assembly direction in world coordinates, as shown in the bottom-left image.}}
    \figlab{method}
\end{figure*}
%%%%%%%%%%%%%%%%%%%%%%%%%%%%%%%%%%

Assembly planning based on \textit{constraints} such as contact between parts has been discussed~\cite{Hirukawa1991,Yoshikawa1991,Hirai1992,Yokokohji1993}. 
Robot task planning based on contact state transitions defined by infinitesimal displacements of the target objects has been extensively discussed~\cite{Hirai1991,Ikeuchi1994,Takamatsu2007,Wake2020}.
They chose a task from several possible transitions of the contact states where the degree of the constraints is increased slightly.

The insertion task (\figref{issue}) is difficult because of the difficulty of the contact state transitions.
In this study, to alleviate such difficulties in the transitions named \textit{constraint state transition difficulty} (CSTD) proposed in~\cite{Yoshikawa1991}, we redesigned the fitness function for the GA~\cite{Tariki2020}.

We used two fitness functions: one to evaluate the insertion condition and another to reduce the CSTD of the sequences. 
As the tradeoff between the two objectives, we need to solve a {\em multiobjective optimization} (MO) problem. 

To minimize production time and cost, Choi~\etal~\cite{Choi2009} applied multicriteria ASG using a given dataset with 19 parts. 
They did not discuss the criteria for reducing the difficulty of the assembly operations or how to extract the necessary data from the models.
We performed the MO using a {\em multiobjective GA} (MOGA)~\cite{Coello2002} to investigate the possibility of finding a Pareto-optimal sequence. 

The ASG for deformable parts is another issue that must be solved.
All the aforementioned methods can only handle rigid parts.
We propose a 3D model based method for obtaining interference-free, insertion, and degree of constraint matrices for deformable parts. 
Deformable objects with a large volume (\eg~seat, cover, and cloth) are beyond the scope of the present study, as each deformable object may require a shape-specific ASG.

Wolter~\etal~\cite{Wolter1996} proposed an operation method for string-like parts (\eg~wires, cables, hoses, and ropes) based on a state representation for part shapes.
To plan a sequence of movement primitives for string-like deformable objects, Takamatsu~\etal~\cite{Takamatsu2006} proposed a knot-state representation for knot-tying.
Dual-armed assembly tasks based on an elastic energy and a collision cost~\cite{Ramirez2018} and step-by-step assembly strategies demonstrated the insertions of ring-shaped deformable objects such as rubber bands~\cite{Kim2020,Ramirez2018} and roller chains~\cite{Tatemura2020}.
By deforming the part model, we determined the interference-free directions and assembly order for string-like and ring-shaped deformable parts.

This study makes four contributions.
(i) We designed a fitness function to generate sequences in which the CSTD is minimized. 
(ii) We developed an MOGA that can find Pareto-optimal sequences.
(iii) We extended the method for extracting two-part relationships for deformable parts.
(iv) To determine the degree of robustness and reproducibility, we extensively evaluated our ASG using eight models made up of rigid and deformable parts.

\section{Assembly Sequence Generation}
%%%%%%%%%%%%%%%%%%%%%%%%%%%%%%%%%%
\begin{figure}[tb]
    \centering
    \includegraphics[width=0.75\linewidth]{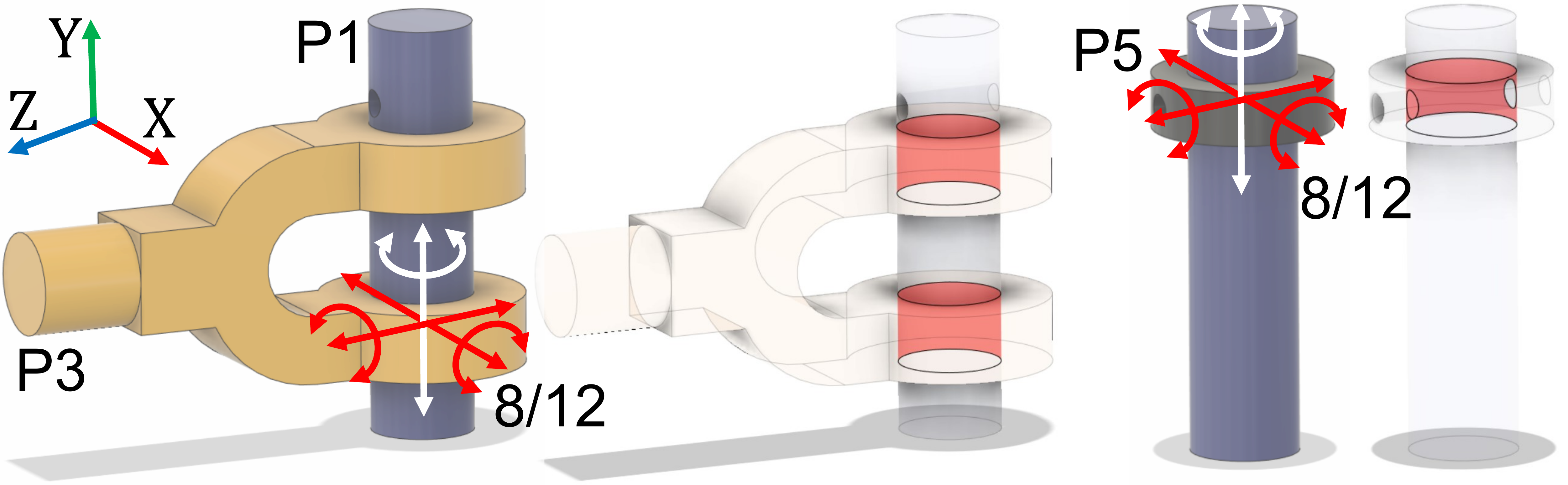}
    \caption{\small{Determination of degrees of constraints. The red shapes and arrows corresponds to the contact surfaces and P1 interference directions. The numbers indicate the degree of constraint per degree of freedom.}}
    \figlab{cstd}
\end{figure}
%%%%%%%%%%%%%%%%%%%%%%%%%%%%%%%%%%
This study is based on the following two assumptions.
(1) We use the same dual-arm robot, mechanical grippers, and assembly jigs for assembly operations as current manufacturing operations.
(2) The proposed algorithm outputs assembly sequences represented as orders of part IDs (\eg~Part 3, Part 17, ...) and the corresponding assembly directions (\eg~$\mathrm{-z}$, $\mathrm{-z}$, ...) in 3-axis coordinates. 

\figref{method} shows the proposed ASG.
First, we extract the parts geometries from the assembled CAD model, then calculate the interference-free, insertion, and proposed degree of constraint matrices. 
Second, the order and assembly direction of the parts are generated using the proposed MOGA.

\subsection{Extraction of Part Relationships between Rigid Parts}
For the proposed ASG, we need the three matrices shown in~\figref{method}. 
In terms of the interference-free~\cite{Tariki2019} and insertion~\cite{Tariki2020} matrices of rigid parts, we extract geometric information from the 3D models using CAD software and calculate them using the method described in~\cite{TarikiAR}.
This section concentrates on describing the CSTD and the method for calculating it.

We calculate the degree of constraint $C(P_i,P_k)$ between parts $P_i$ and $P_k$. 
If there is no contact between the parts, this value is set to 0. 
According to Yoshikawa~\etal~\cite{Yoshikawa1991}, the degree of constraint is defined as:
\begin{equation}\forlab{cik}
   C(P_i,P_k)
   =
   12 - \sum_{j=1}^{12} F_j(P_i,P_k)~\in \{0,1,..,11\},
\end{equation}
where $F_j(P_i,P_k)~(j=1,2,...,12)$ indicates constraint-free information for 12 directions of translational and rotational displacements $\pm{{\mathrm{x}}},\pm{{\mathrm{y}}},\pm{{\mathrm{z}}}$ and $\pm{\Theta_{\mathrm{x}}},\pm{\Theta_{\mathrm{y}}},\pm{\Theta_{\mathrm{z}}}$ of the X, Y, and Z axes shown in~\figref{cstd}. 
This value is set to 1 if the parts do not interfere with each other after an infinitesimal displacement.
Otherwise the value is 0. 

We note that moving P1 in the $\mathrm{+x}$ direction and moving P2 in the $\mathrm{-x}$ direction are the same in terms of the relationship between P1 and P2, thus, $F_1(P_i,P_k)=F_2(P_k,P_i)$. 
All other directions have the same relationship.
Therefore, to reduce the time to calculate the function $\bm{F}$, the interference-free information for the negative directions of all axes are calculated as the transpose of the matrix on the positive direction of each corresponding axis. 
Finally, the matrix for the degree of constraint $\bm{C}$ is computed using \forref{cik} as an element. 
Because $\bm{C}$ is symmetrical, we calculate only the upper triangular component and calculate the other elements based on the relationship $C(P_i,P_k)=C(P_k,P_i)$.

Given that the assembly order $P_{O_1},P_{O_2},..,P_{O_{k}}$, the maximum CSTD $H$ is calculated as:
\begin{equation}\forlab{hs}
   H
   \coloneqq
   \max_{k \in \{2,3,..,\eta\}} \sum_{i=1}^{k-1} C(P_{O_i},P_{O_k}),
\end{equation}
where $\sum_{i=1}^{k-1} C(P_{O_i},P_{O_k})$ shows the CSTD in the assembly of the $k$-th part $P_{O_k}$ and the other assembled parts $P_{O_1},P_{O_2},..,P_{O_{k-1}}$.

To calculate the CSTD, the constraint-free information of an arbitrary part is determined by investigating whether a part interferes with other parts, as illustrated in~\figref{cstd}. 
In the figure, the investigated target part is displaced in six positive and negative directions along the X, Y, and Z axes and rotated around the X, Y, and Z axes.
The origin of the coordinate system is automatically determined as the center of gravity of the shape composed of a contact surface (constraint surface) between the two parts.
The Z-axial positive direction of the coordinate system is determined as the direction vertically upward in a stable pose of the product with the widest bottom surface to place on a plane. 
If multiple contact surfaces are found, one of them is randomly selected.
The positive directions of the X and Y axes are determined in the directions of the world coordinate system of the model, and only the rotation center is set by the center of gravity.
\figref{cstd} shows the determined axes on assembled parts in a model.

\subsection{Extraction of Part Relationships for Deformable Parts}
%%%%%%%%%%%%%%%%%%%%%%%%%%%%%%%%%%
\begin{figure}[tb]
  \begin{minipage}[tb]{0.47\linewidth}
    \centering
    \includegraphics[width=0.7\linewidth]{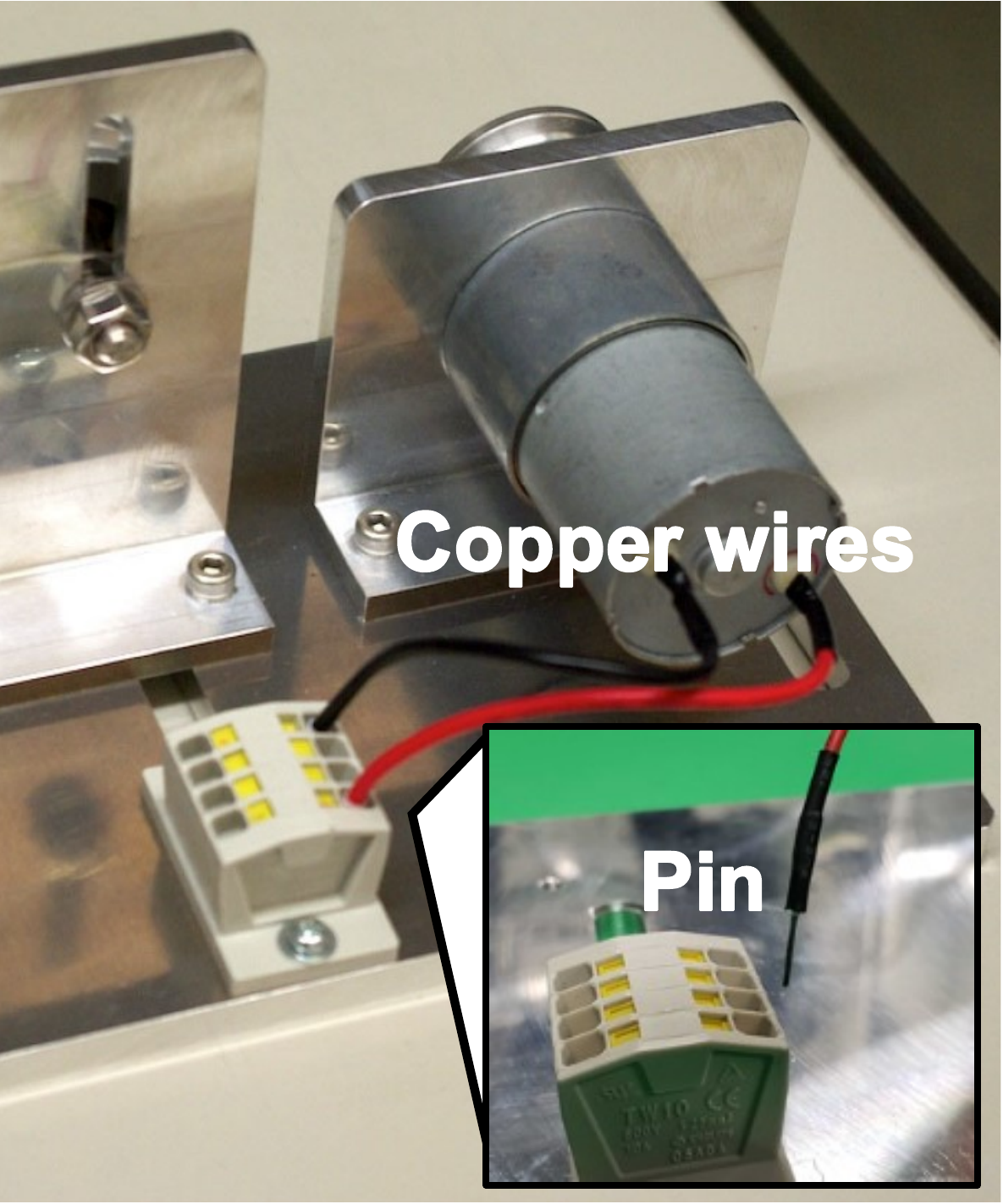}
    \subcaption{\small{String-like~\cite{WRS2020}}}
  \end{minipage}
  \begin{minipage}[tb]{0.52\linewidth}
    \centering
    \includegraphics[width=0.6\linewidth]{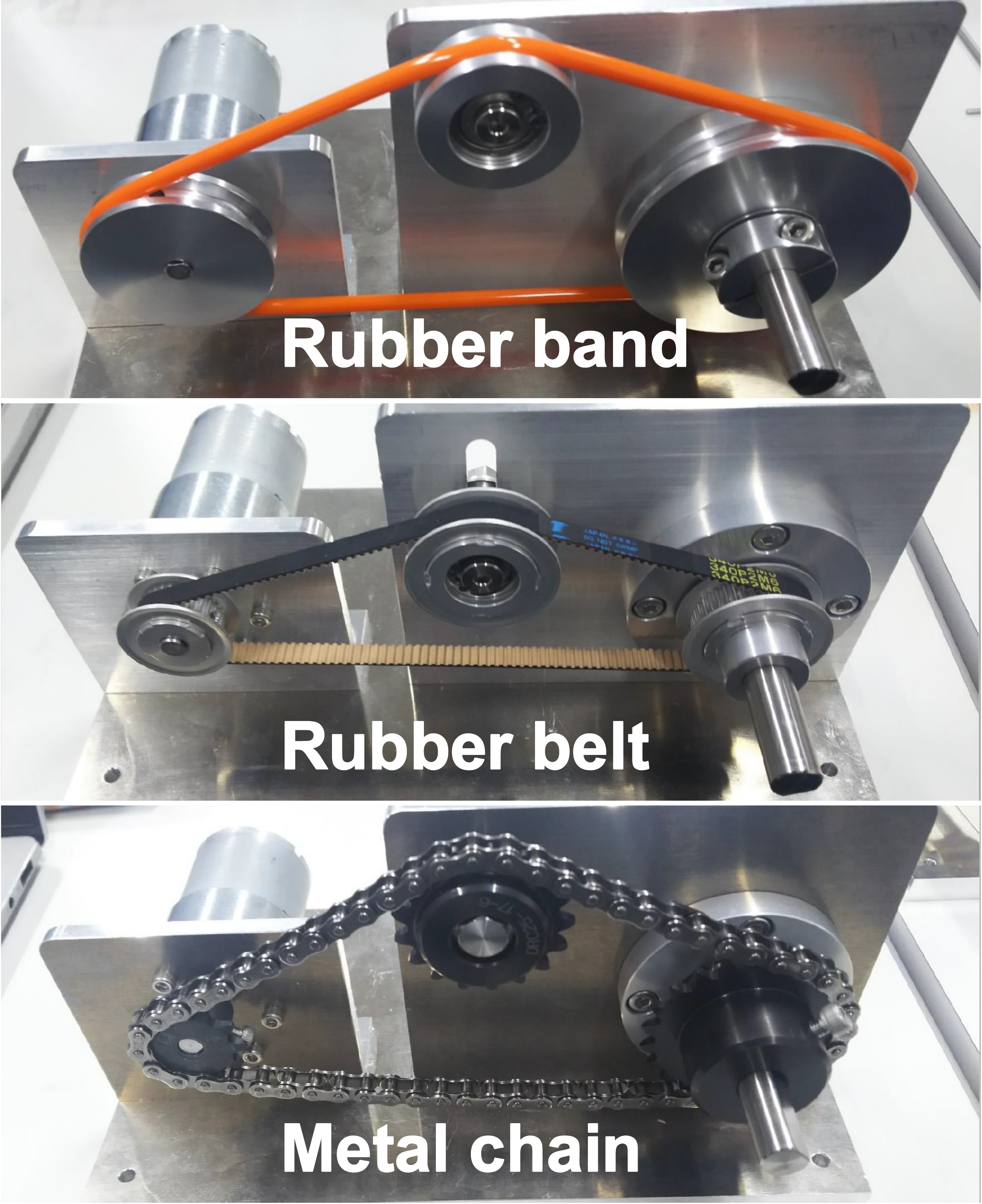}
    \subcaption{\small{Ring-shaped~\cite{WRS2018}}}
  \end{minipage}
  \caption{\small{Two types of deformable parts.}}
  \figlab{deform}
\end{figure}
%%%%%%%%%%%%%%%%%%%%%%%%%%%%%%%%%%
\begin{figure}[tb]
  \begin{minipage}[tb]{\linewidth}
    \centering
    \includegraphics[width=0.79\linewidth]{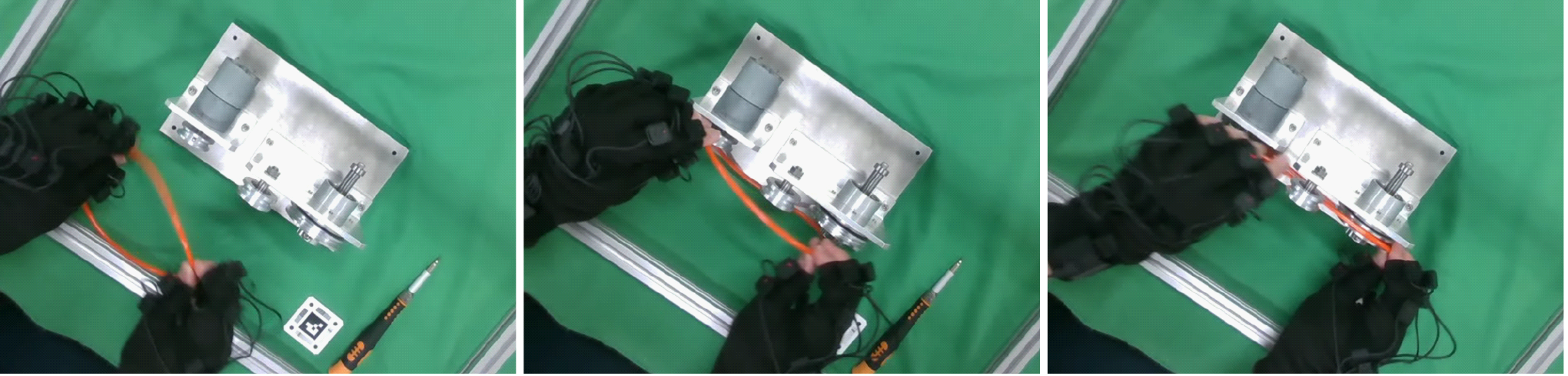}
    \subcaption{\small{Avoiding interference by deforming a rubber band}}
  \end{minipage}
  \begin{minipage}[tb]{\linewidth}
    \centering
    \vspace{2pt}
    \includegraphics[width=0.78\linewidth]{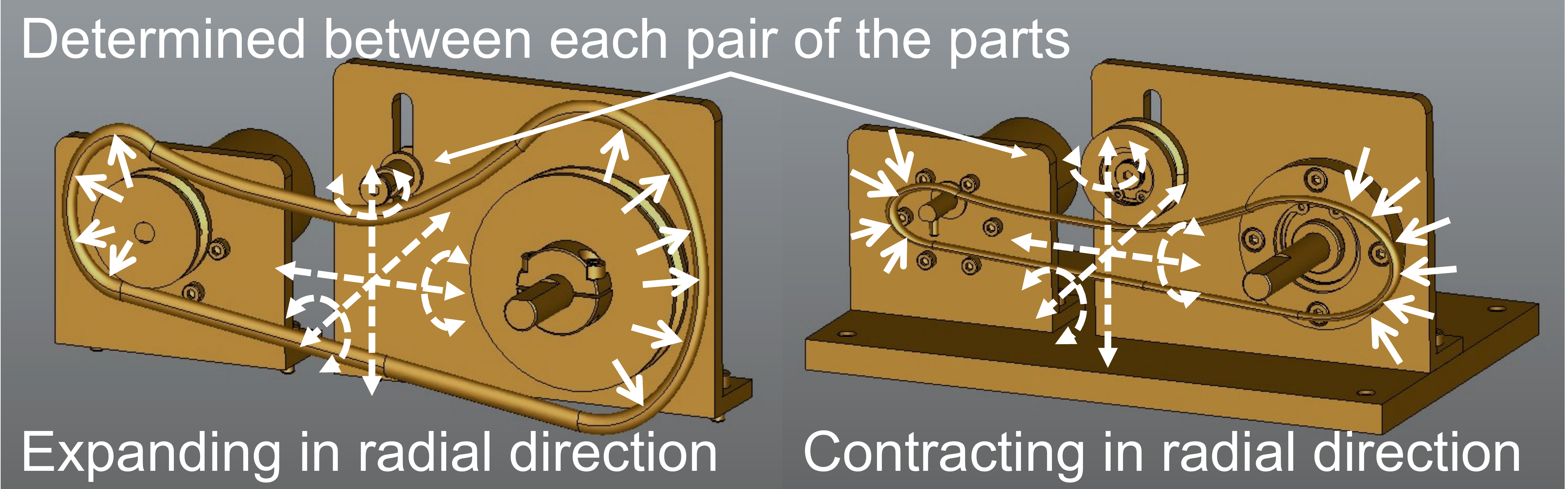}
    \subcaption{\small{Determination of interference}}
  \end{minipage}
  \caption{\small{Interference determination for a deformable part. The model is deformed in the radial direction (b) as a human would do (a).}}
  \figlab{judge-deform}
\end{figure}
%%%%%%%%%%%%%%%%%%%%%%%%%%%%%%%%%%
\figref{deform} shows string-like deformable parts that will be used in the assembly challenge of WRS2020~\cite{WRS2020} and the ring-shaped deformable parts used in the assembly challenge of WRS2018~\cite{WRS2018}.
This study concentrates on string-like deformable parts, such as the wire with a rigid pin shown in~\figref{deform}(a) and ring-shaped deformable parts such as the rubber band, rubber belt, and metal chain shown in~\figref{deform}(b).

\subsubsection{String-like parts}
String-like deformable parts, such as a cable with a plug or a wire with pins, are often combined with a rigid body attached to the tip as shown in~\figref{deform}(a).
String-like deformable parts, such as connectors, cables and wires, appear frequently in products. Both the plug and pin are attached for inserting into or connecting to others such as a socket and a hole. 
Thus, if the string-like deformable object has a rigid part connected to others, the two-part relationships between the rigid part and others must be investigated.

For example, the vertices of string-like parts and the corresponding inserted part are recognized, then the system calculates the interference-free, insertion, and degree of constraint matrices between them in the same way as for rigid parts.
This implies that the deformable region for a string-like deformable object can be disregarded. 
Entanglement with other parts needs to be considered~\cite{Sanchez2020}; however, this is beyond the scope of the present study.

\subsubsection{Ring-shaped parts}
We describe a method for extracting the constraint-free information for a rubber band as an example of ring-shaped deformable parts. 
We assumed that the part deformability can be determined from the part name.

For example, the rubber band shown in~\figref{judge-deform} transmits the rotation of the motor shaft to another pulley. 
The rubber band must be stretched and retracted in the radial direction when attached to a pulley groove in the assembly as a human would do.

By expanding or contracting the model in the radial direction, the constraint-free information of its deformed shape is extracted, as shown in~\figref{judge-deform} (b). 
We changed the scaling factor for the deformation of the mesh part model in CAD.
If any one of the extracted constraint-free information with 12 directions becomes 1, the scaling factor is adopted. 
The three matrices for the proposed ASG are obtained in the same way as for the rigid parts. 
The elements of the insertion matrix for the ring-shaped parts are set to zero.

\subsection{Optimization using MOGA}
To solve the MO problem, we built an algorithm based on NSGA-II~\cite{Deb2002}, an MOGA that provides high search performance for 2--3 objective MOPs.
\figref{nsga2} shows the proposed algorithm. 
We designed the fitness function to evaluate the insertion condition and CSTD between the parts.
The blue part of~\figref{nsga2} is detailed in~\cite{TarikiAR} and includes chromosome coding, chromosome initialization, and genetic operation.
%%%%%%%%%%%%%%%%%%%%%%%%%%%%%%%%%%
\begin{figure}[tb]
    \centering
    \includegraphics[width=0.75\linewidth]{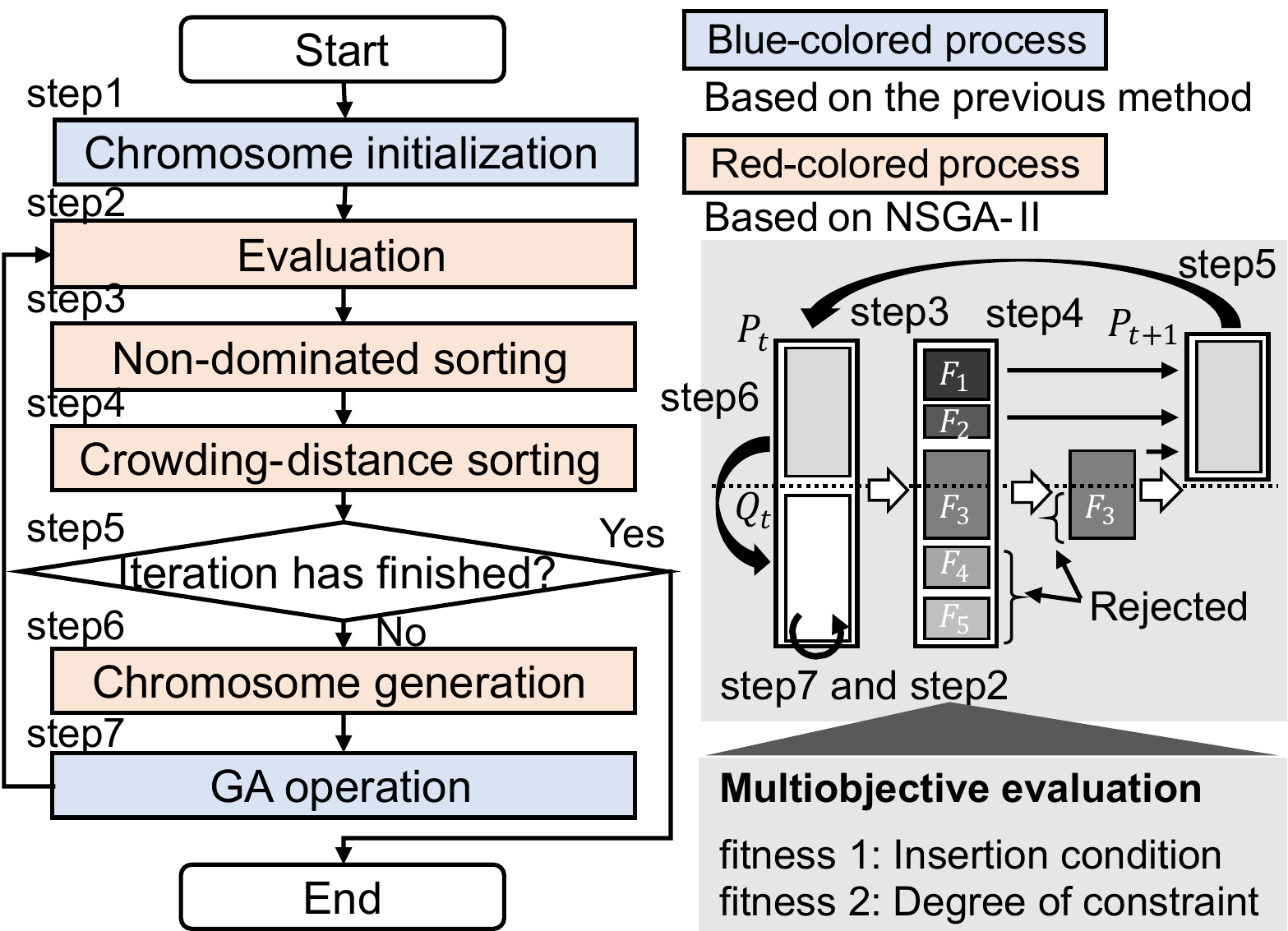}
    \caption{\small{Assembly sequence optimization. The blue and red blocks are based on the previous method~\cite{TarikiAR} and NSGA-II~\cite{Deb2002}, respectively.}}
    \figlab{nsga2}
\end{figure}
%%%%%%%%%%%%%%%%%%%%%%%%%%%%%%%%%%
\begin{table}[tb]
    \centering
        \caption{\small{GA parameters used in our experiments.}} \tablab{results}
        \begin{tabular}{lr} \toprule
            \begin{tabular}{c}Parameter\end{tabular}
            &\begin{tabular}{c}Value\end{tabular} \\ \midrule
            Number of chromosomes & $\eta$ \rule[-1mm]{0mm}{4mm} \\
            Crossover rate & 0.2 \rule[-1mm]{0mm}{4mm} \\
            Mutation rate & 0.1 \rule[-1mm]{0mm}{4mm} \\
            Cut-and-paste rate & 0.35 \rule[-1mm]{0mm}{4mm} \\
            Break-and-join rate & 0.35 \rule[-1mm]{0mm}{4mm} \\
            Generation update & 100 \rule[-1mm]{0mm}{4mm} \\
            Number of iterations & 10 \rule[-1mm]{0mm}{4mm} \\ \bottomrule
        \end{tabular}
\end{table}
%%%%%%%%%%%%%%%%%%%%%%%%%%%%%%%%%%
The fitness function that satisfies the insertion condition (hereinafter referred to as Fitness~1)~\cite{Tariki2020} is
\begin{equation}\forlab{fi}
  f_i(s)
  \coloneqq
 \left\{
  \begin{array}{ll}
     2\eta+\alpha(s)-\beta(s)-r(s)  &  \text{feasible} \\
     \eta/2                         &  \text{infeasible}
    \end{array},
 \right.
\end{equation}
where $\eta$ is the number of parts, $s$ indicates the sequence, $\alpha(s)$ and $\beta(s)$ are parameters related to the insertion condition, and $r(s)$ is the number of changes in the assembly direction. 

The fitness function for the CSTD is designed such that, if the assembly is infeasible, the evaluation is the lowest; otherwise, it is designed such that the sequence with the lowest CSTD receives the highest evaluation. 
Minimizing the CSTD must be solved for each part assembly based on the fitness function (hereinafter referred to as Fitness~2) calculated as:
\begin{equation}\forlab{fc}
  f_c
  \coloneqq
 \left\{
  \begin{array}{ll}
     12(\eta-1)-H  &  \text{feasible} \\
     0      &  \text{infeasible}
    \end{array}
 \right..
\end{equation}
This value is 0 for infeasible assembly. 
The feasibility is determined using the method devised by Smith~\etal~\cite{Smith2001}.
In \forref{fc}, $H$ is the maximum CSTD.
According to the definition, the maximum constraint for two parts is 12; therefore, $H$ in \forref{fc} is less than $12(\eta-1)$.

\section{Experiments}
%%%%%%%%%%%%%%%%%%%%%%%%%%%%%%%%%%
\begin{figure}[tb]
  \centering
  \includegraphics[width=0.55\linewidth]{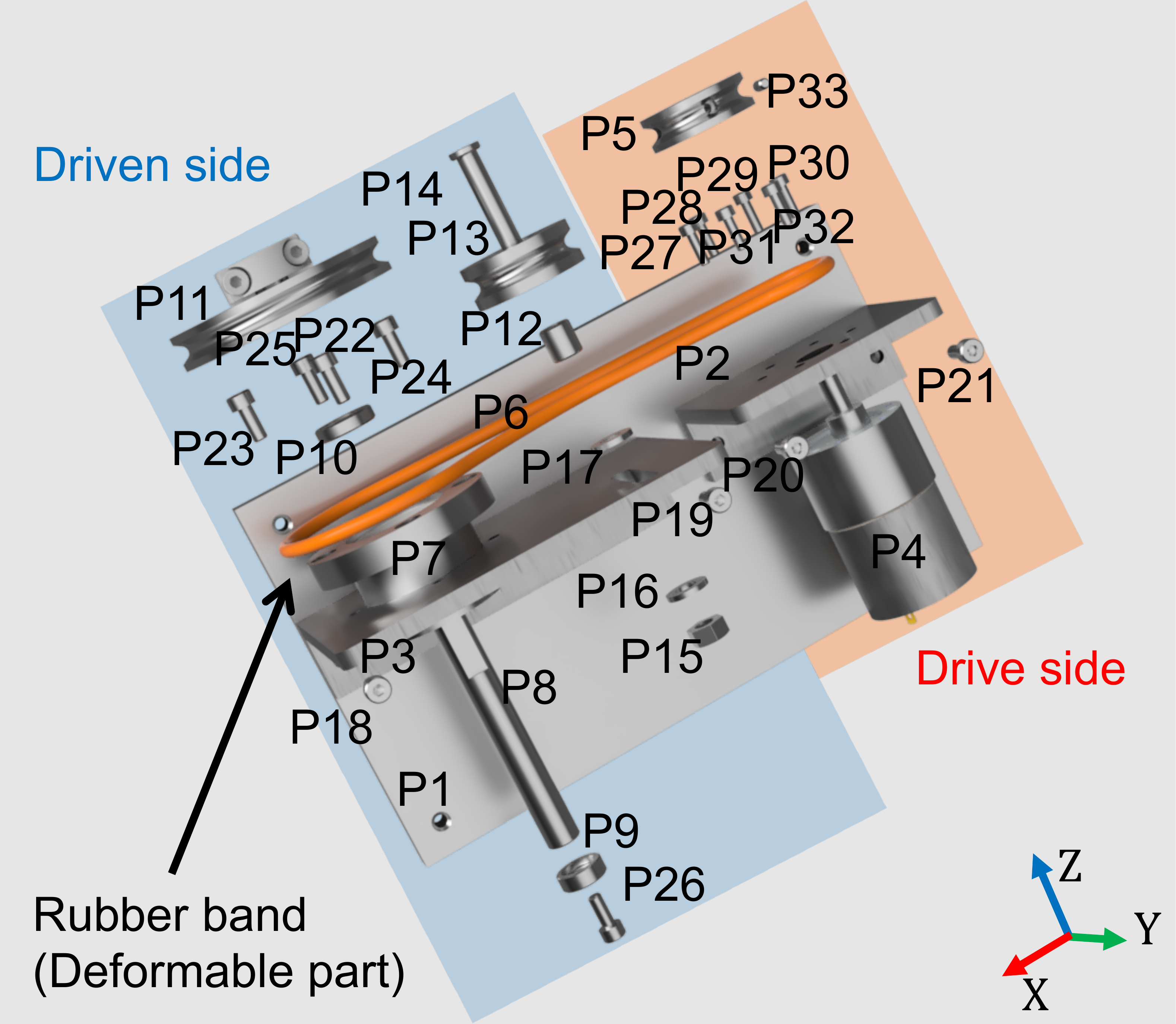}
  \caption{\small{Rubber-band drive unit (\#2) used in Case Study~2.}}
  \figlab{case2}
\end{figure}
%%%%%%%%%%%%%%%%%%%%%%%%%%%%%%%%%%
\begin{figure}[tb]
  \begin{minipage}[tb]{0.49\linewidth}
    \centering
    \includegraphics[width=0.71\linewidth]{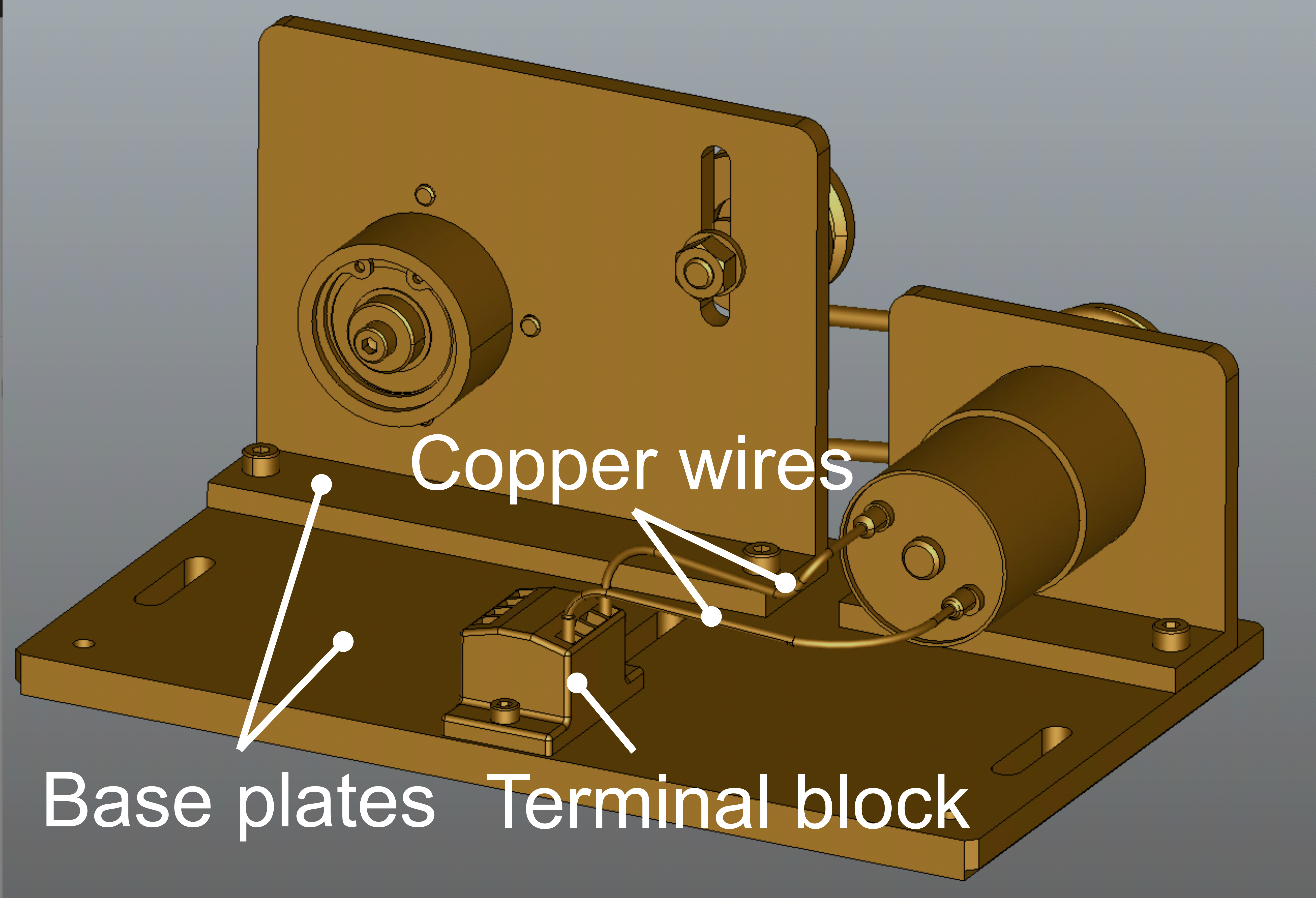}
    \subcaption{\small{Rubber band-drive unit (\#3)}}
  \end{minipage}
  \begin{minipage}[tb]{0.49\linewidth}
    \centering
    \includegraphics[width=0.87\linewidth]{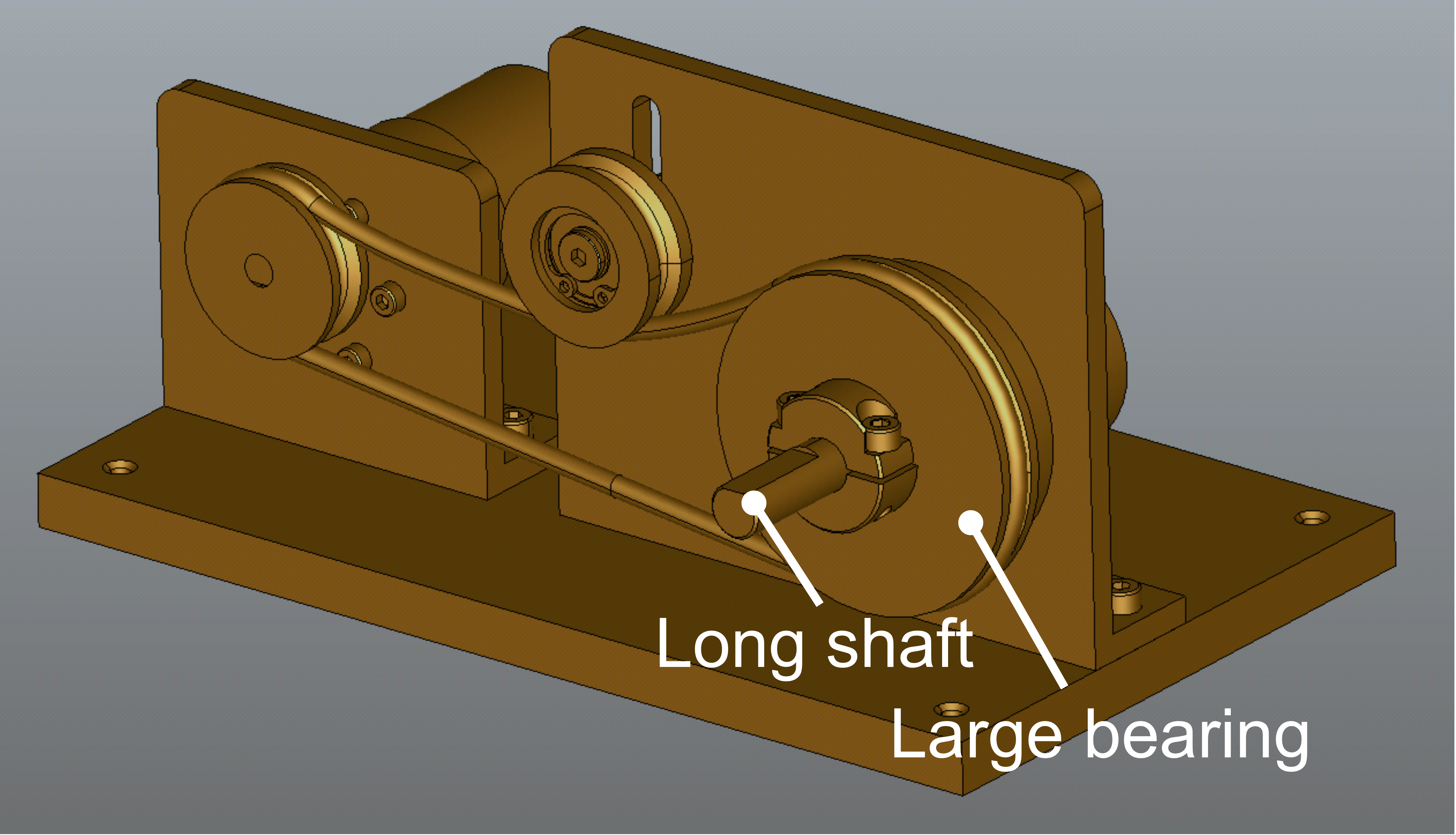}
    \subcaption{\small{Rubber-band drive unit (\#4)}}
  \end{minipage}
  \begin{minipage}[tb]{0.49\linewidth}
    \centering
    \includegraphics[width=0.87\linewidth]{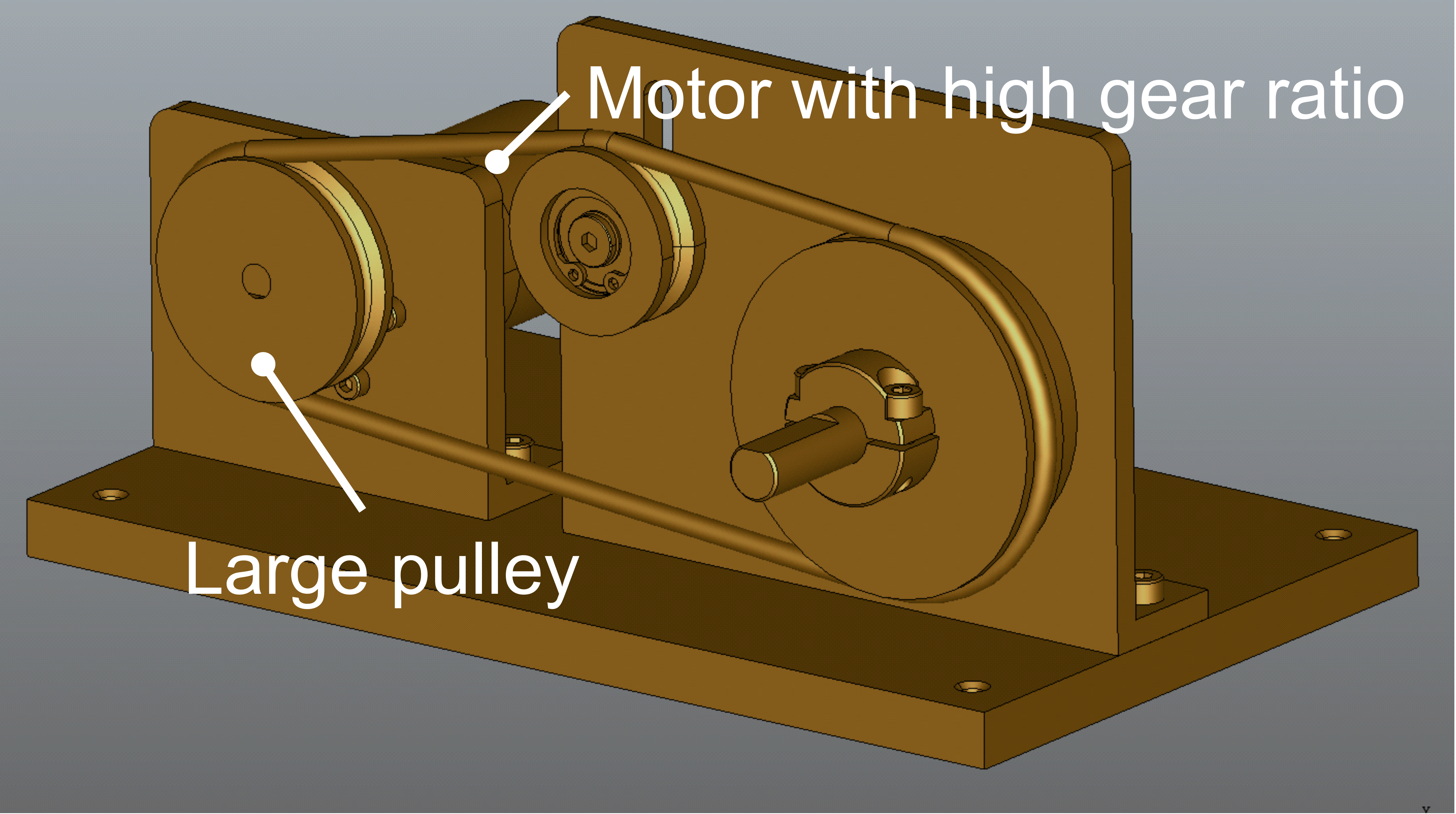}
    \subcaption{\small{Rubber-band drive unit (\#5)}}
  \end{minipage}
  \begin{minipage}[tb]{0.49\linewidth}
    \centering
    \includegraphics[width=0.87\linewidth]{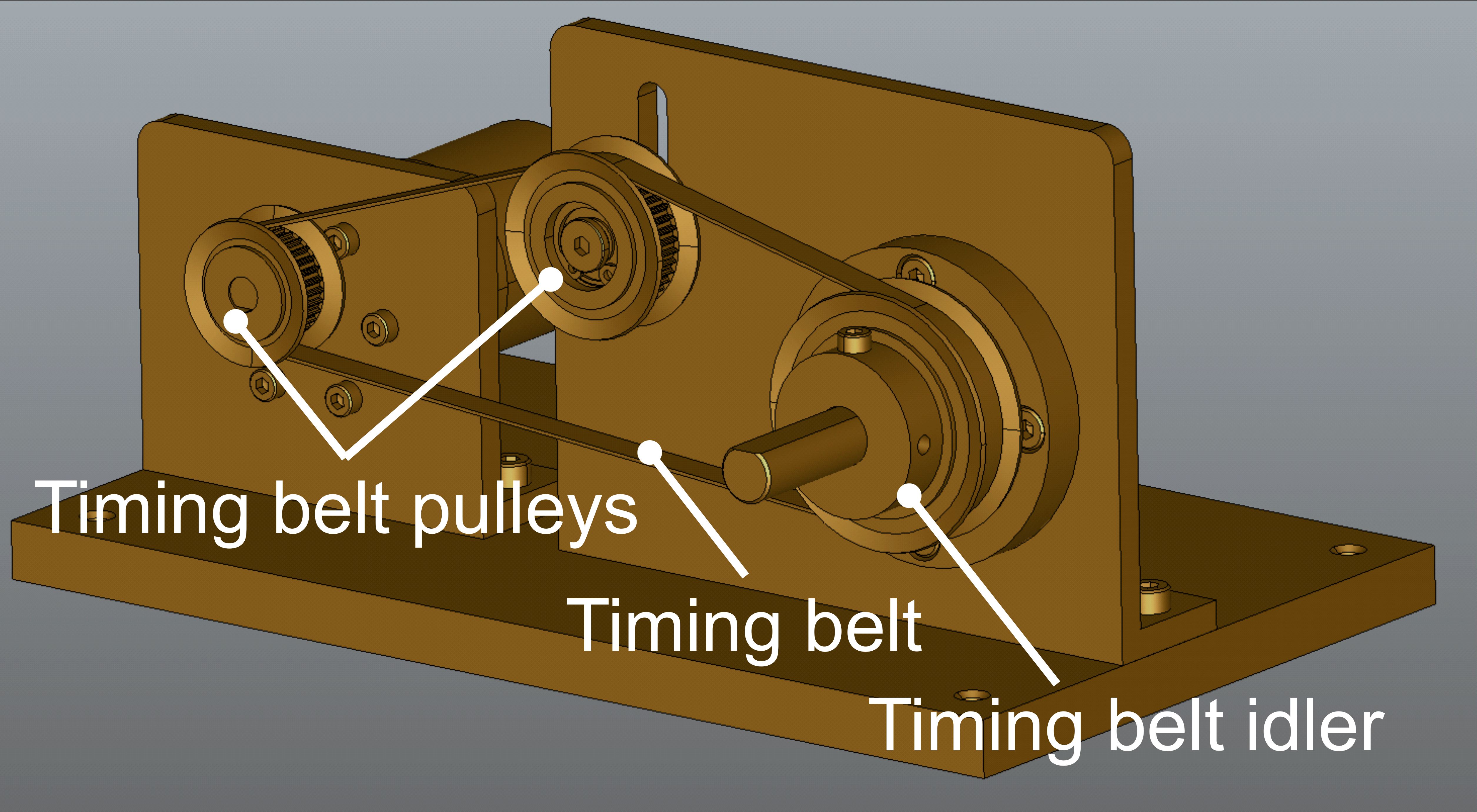}
    \subcaption{\small{Rubber-belt drive unit (\#6)}}
  \end{minipage}
  \begin{minipage}[tb]{0.49\linewidth}
    \centering
    \includegraphics[width=0.87\linewidth]{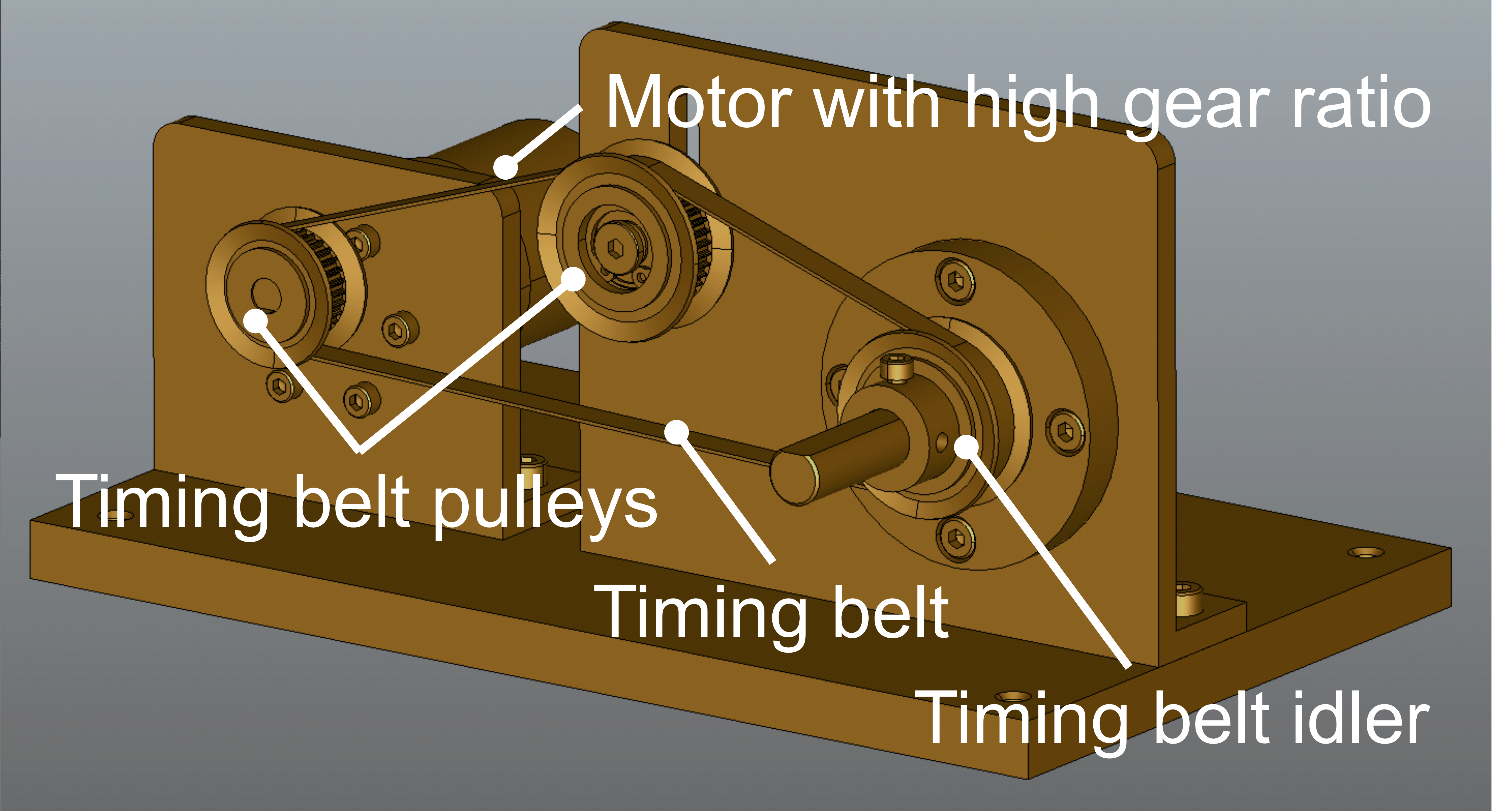}
    \subcaption{\small{Rubber-belt drive unit (\#7)}}
  \end{minipage}
  \begin{minipage}[tb]{0.49\linewidth}
    \centering
    \includegraphics[width=0.8\linewidth]{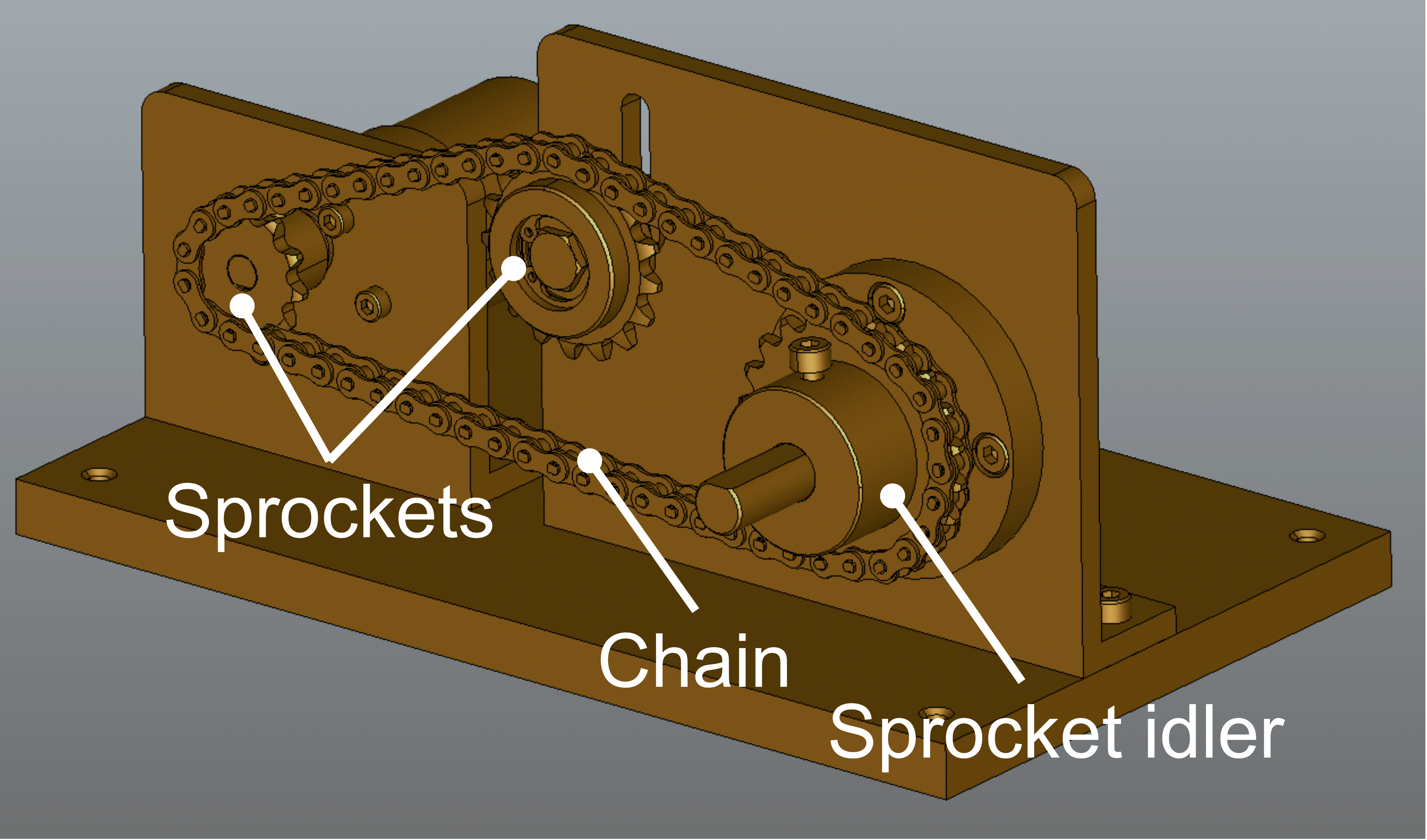}
    \subcaption{\small{Chain-drive unit (\#8)}}
  \end{minipage}
  \caption{\small{Models used in Case Study~3. The six models are revised from \#2 shown in~\figref{case2}. The replaced parts are indicated in each figure.}}
  \figlab{cs3-models}
\end{figure}
%%%%%%%%%%%%%%%%%%%%%%%%%%%%%%%%%%
\begin{figure}[tb]
  \centering
  \includegraphics[width=0.55\linewidth]{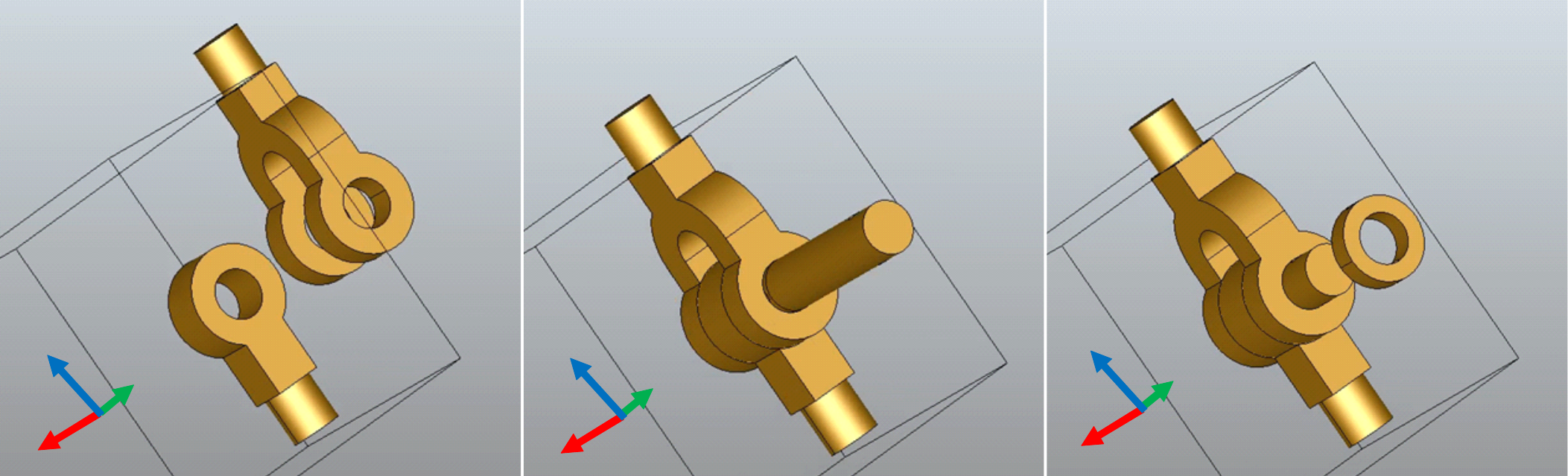}
  \caption{\small{Sequence generated in Case Study~1.}}
  \figlab{result_cs1}
\end{figure}
%%%%%%%%%%%%%%%%%%%%%%%%%%%%%%%%%%
\begin{figure*}[tb]
  \begin{minipage}[tb]{0.325\linewidth}
    \centering
    \vspace{3mm}
    \includegraphics[width=0.88\linewidth]{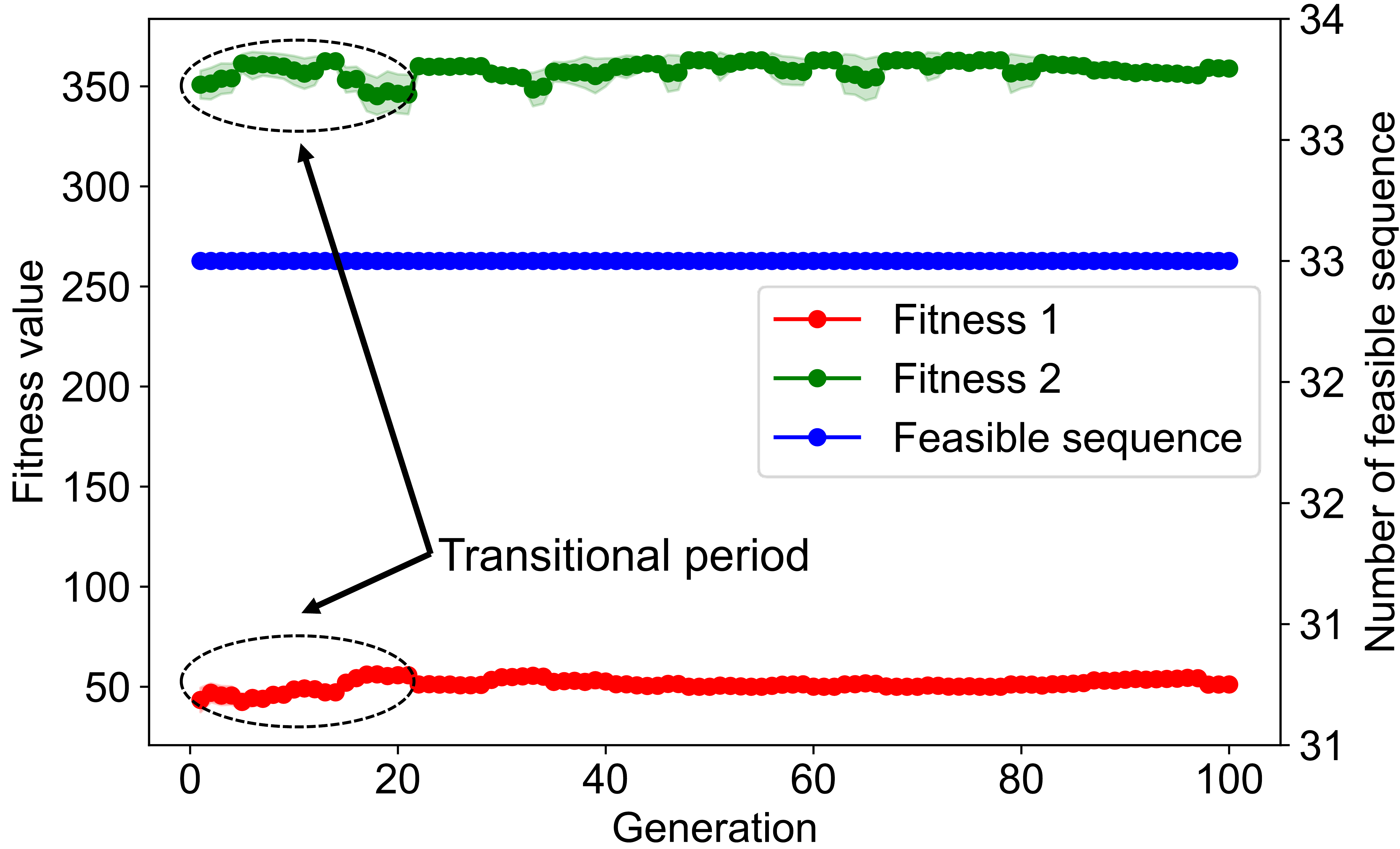}
    \vspace{1mm}
    \subcaption{\small{Convergence curve in one optimization}}
  \end{minipage}
  \begin{minipage}[tb]{0.325\linewidth}
    \centering
    \vspace{2mm}
    \includegraphics[width=0.88\linewidth]{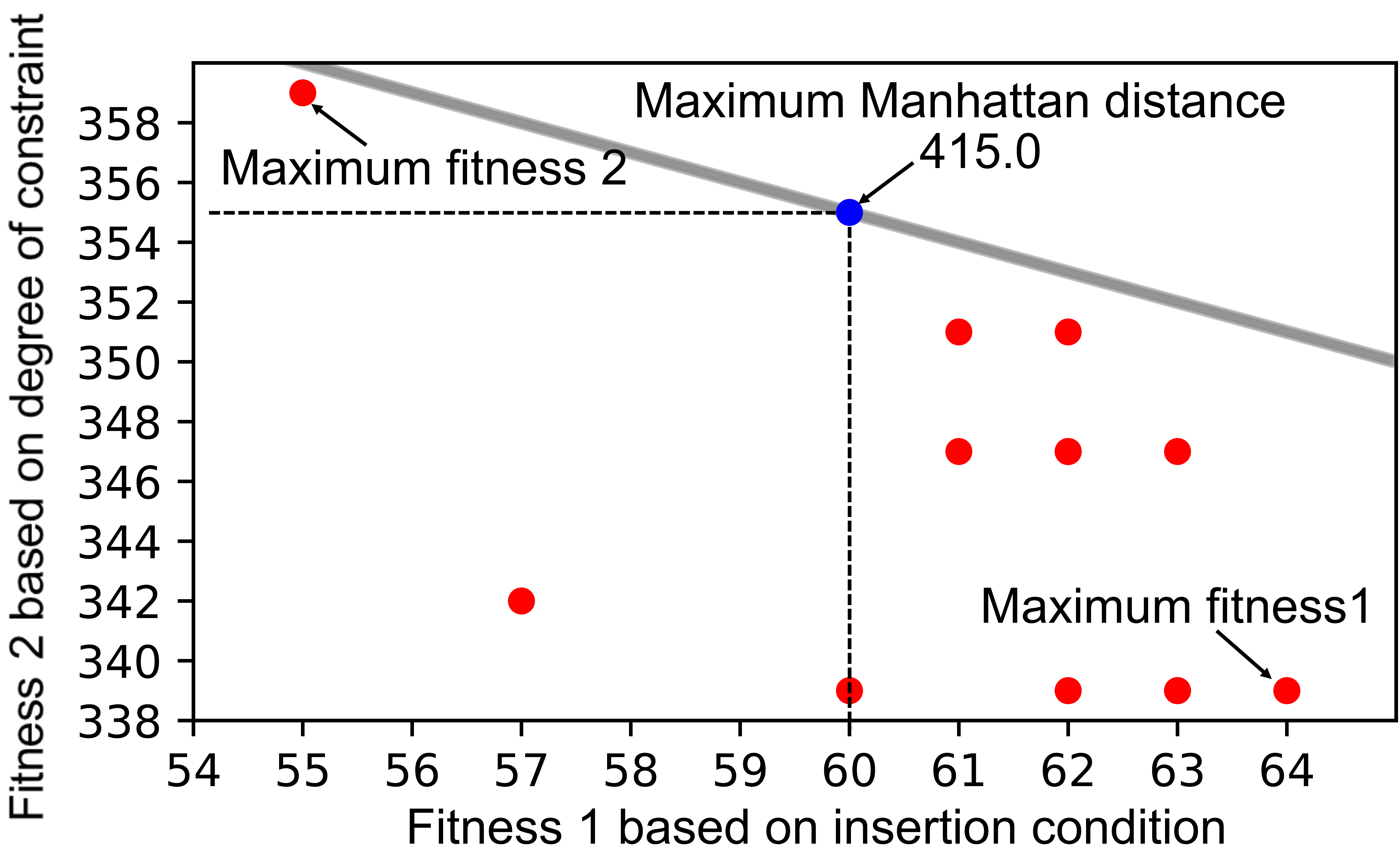}
    \vspace{1.5mm}
    \subcaption{\small{Fitness values of generated sequences}}
  \end{minipage}
  \begin{minipage}[tb]{0.325\linewidth}
    \centering
    \includegraphics[width=\linewidth]{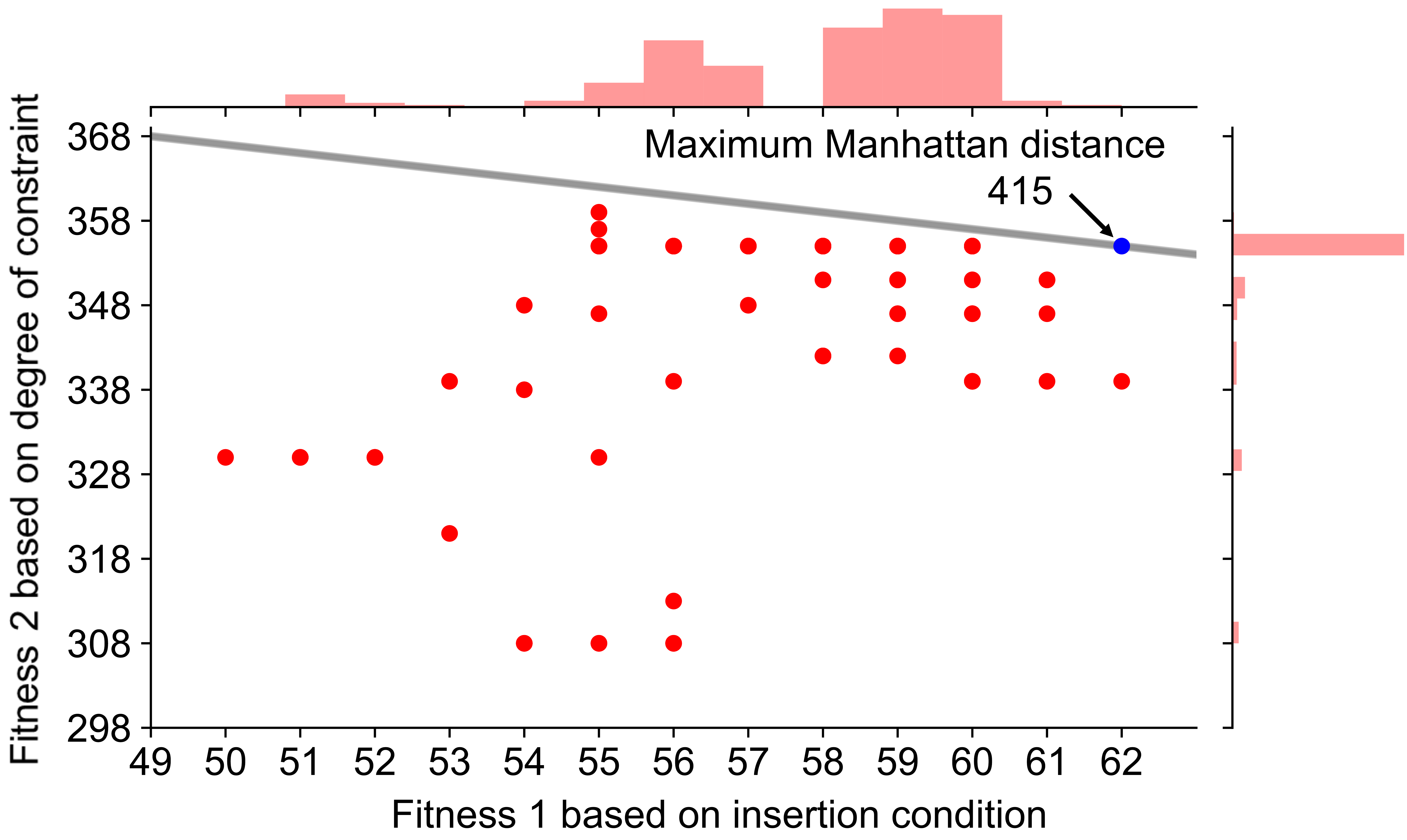}
    \subcaption{\small{After reordering of generated sequence}}
  \end{minipage}
  \caption{\small{Results of Case Study~2. (a) Progress of fitness values and number of feasible sequences identified during one optimization. (b) Fitness values in 100-th generation. (c) Fitness values after reordering of the generated sequence shown in (b) as a blue dot.}}
  \figlab{sim-res-cs2}
\end{figure*}
%%%%%%%%%%%%%%%%%%%%%%%%%%%%%%%%%%
\begin{figure*}[tb]
    \centering
    \includegraphics[width=0.9\linewidth]{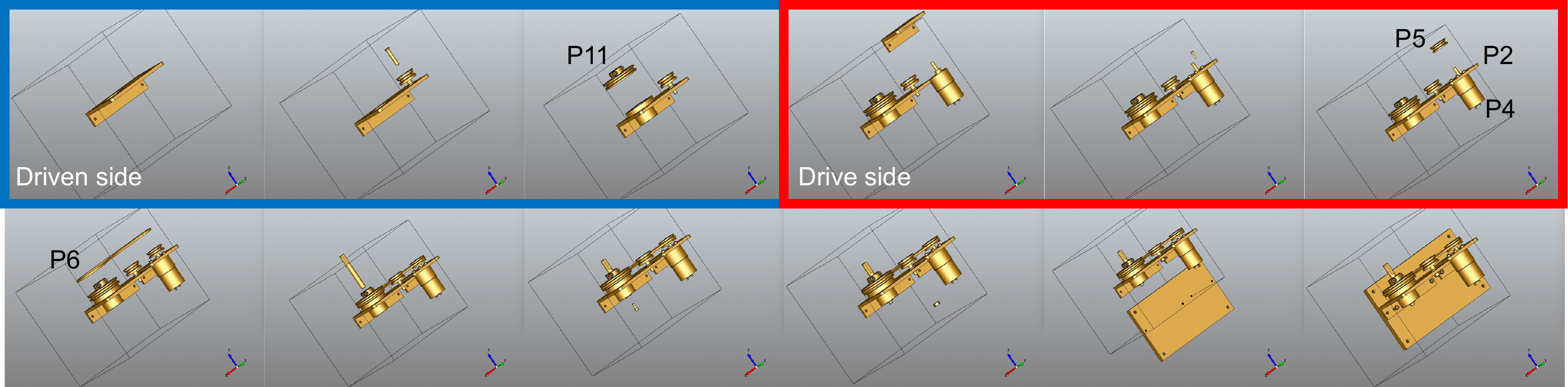}
    \caption{\small{Sequence generated in Case Study~2. The order (assembly direction) is: 3 (-z), 17 (-z), 12 (-z), 13 (-z), 14 (-z), 7 (-z), 23 (-z), 25 (-z), 24 (-z), 22 (-z), 10 (-z), 11 (-z), 4 (-z), 2 (-z), 27 (-z), 29 (-z), 30 (-z), 31 (-z), 32 (-z), 28 (-z), 5 (-z), 33 (-y), 6 (-z), 8 (-z), 9 (+z), 26 (+z), 16 (+z), 15 (+z), 1 (+z), 19 (-y), 21 (-y), 20 (-y), 18 (-y).}}
    \figlab{sequence_cs2}
\end{figure*}
%%%%%%%%%%%%%%%%%%%%%%%%%%%%%%%%%%
\subsection{Outline}
We conducted three MOGA case studies with the parameters listed in~\tabref{results}.
Case Study~1 used the model shown in~\figref{issue} to confirm whether the aformentioned problem can be solved. 

Case Study~2 used the model of a rubber-band drive unit (\figref{case2}) consisting of 33 parts used for an assembly challenge~\cite{Yokokohji2019}. 
We investigated the possibility of applying the ASG for many parts, including a deformable part. 

Case Study~3 was conducted to verify whether the proposed method can generate sequences for several models where the types of some parts differ slightly.
\figref{cs3-models} shows the eight models used for Case Study~3: the two models used in Case Study~1 (\#1) and 2 (\#2), a model that includes copper wires with pins inserted into a terminal block extending from the rubber-band drive unit used in Case Study~2 (\#3), two rubber-band drive units which are different from the model used in Case Study~2 (\#4 and \#5), two rubber-belt drive units  (\#6 and \#7), and a chain-drive unit (\#8) used in the assembly challenge~\cite{WRS2018,WRS2020}.
Furthermore, using three models \#1, \#2, and \#3, we evaluated the reproducibility of the ASG.

\subsection{Case Study~1}
\figref{result_cs1} shows the final assembly sequence with the highest sum of the fitness values of \forref{fi} and \forref{fc} among the generated sequences in 10 trials. 
The sequence shown in the left-hand side in~\figref{issue}, depicting the simultaneous occurrence of contacts was removed and the assembly sequence with a low CSTD was generated.

\figref{issue} (left) shows that when P1 is inserted, constraints occur simultaneously on P2, P3, and P5, and the CSTD is 24 (= 8 + 8 + 8). 
In contrast, as shown in~\figref{result_cs1}, when P1 is inserted, constraints occur with both P3 and P5, and the CSTD is 16 (= 8 + 8). In the insertion of P5, the constraints with only P1 and P3 are 13 (= 8 + 5).
In both these cases, the CSTD is less than 24. The assembly of the other parts also shows a CSTD of less than 16; thus, the maximum value of the CSTD could be reduced from 24 to 16.

\subsection{Case Study~2}
%%%%%%%%%%%%%%%%%%%%%%%%%%%%%%%%%%
\begin{table}[tb]
    \centering
    \caption{\small{Computation times for MO using two CPUs in Case Study~2. Each item shows the mean plus-minus twice the standard deviation of times for 10 trials. The feasible sequence rate for the 10 sequences is 100\% for all optimizations.}} \tablab{result_cs2}
    \begin{tabular}{crr} \toprule
        & \multicolumn{2}{c}{Runtime of optimization [h]} \\
        \begin{tabular}{c}Generation update\end{tabular}
        & \begin{tabular}{c}Core i7-3520M$^{\rm *a}$\end{tabular} & \begin{tabular}{c}Core i9-9900KS$^{\rm *b}$\end{tabular} \\ \midrule
        1 & $0.901\pm{0.100}$ & $0.336\pm{0.0360}$ \rule[-1mm]{0mm}{4mm} \\
        5 & $1.53\pm{0.123}$ & $0.572\pm{0.0492}$ \rule[-1mm]{0mm}{4mm} \\
        10 & $2.31\pm{0.202}$ & $0.848\pm{0.0416}$ \rule[-1mm]{0mm}{4mm} \\
        20 & $3.77\pm{0.195}$ & $1.39\pm{0.0726}$ \rule[-1mm]{0mm}{4mm} \\
        \bottomrule
    \end{tabular}
    \begin{tablenotes}
    \item[a]\footnotesize{$^{\rm *a}$Intel Core i7-3520M CPU@2.90GHz}
    \item[b]\footnotesize{$^{\rm *b}$Intel Core i9-9900KS CPU@4.00GHz}
    \end{tablenotes}
\end{table}
%%%%%%%%%%%%%%%%%%%%%%%%%%%%%%%%%%
\figref{sim-res-cs2}(a) shows the convergence curve for the MOGA. 
The two curves show the average fitness values of all the chromosomes of each generation. 
They are calculated using Fitness~1 (red curve) and Fitness~2 (green curve). 
The number of interference-free sequences remained at 33, indicating that 100\% of the generated sequences are feasible. 
This indicates that the values may have converged to quasi-optimal values during the first generation update. An unsteady variation is observed in the evaluated values until near the 20th generation update after which the fitness values of the generated sequences are stable and produce high fitness values. 

In this study, the number of generation updates was 100; however, as shown in~\tabref{result_cs2}, even when the number was the small (such as 1 or 5), the generated sequence was still feasible.
There is room to adjust the number of generation updates to reduce the time required for the MO.

\figref{sequence_cs2} shows the generated sequence with the highest sum of the fitness values depicted as the blue dot in \figref{sim-res-cs2}(b). 
Considering only the insertion condition, it would be reasonable to assemble P5 and P2 before P4. 
However, the CSTD for the insertion sequence of P4 into P5 and P2 is high. 
In the generated sequence, P5 is assembled last, thus the CSTD in the assembly of P2, P4, and P5 was reduced.

\figref{sim-res-cs2}(b) shows the two fitness values for 33 generated sequences ($=\eta$). 
Fitness~1 is over $16.5=\eta/2$ and Fitness~2 is over 0 (these values implies infeasible) in all the generated sequences, and an interference-free sequence was generated even in the assembly for deformable parts. 
The solution near the blue dot in~\figref{sim-res-cs2}(b), where the sum of both fitness values is a maximum, can be a Pareto-optimal sequence. 

To verify the Pareto optimality of the solution with the best fitness value, we investigated whether the other fitness value increases or when the order of one part is changed.
In other words, because finding the optimal solution is NP-complete problem, in this experiment, we show that the solutions in the neighborhood is worse than our final solutions (the sequence with the highest sum of the fitness values).

\figref{sim-res-cs2}(c) shows fitness values of the sequences generated by reordering one part, thus the number of sequences simulated is $1024~(=(\eta-1)^2)$.
The number of feasible sequences is 40.3\% ($=413/1024$).
We confirmed that no sequence obtained by reordering increased both fitness values over the best solution shown as the blue dot in~\figref{sim-res-cs2}(b).
Therefore, the generated sequence may satisfy Pareto optimality.

\subsection{Case Study~3}
The objective of this case study was to confirm the robustness and reproducibility of the proposed ASG.
First, we calculated the interference-free, insertion, and degree of constraint matrices for the eight models.
In \figref{cs3-models}, different assembly parts are written in letters inside each model image (\#4$\sim$\#8).
Because the models \#4$\sim$\#8 have a parts structure similar to model \#2, the two-part relationships were extracted successfully.
In the case of \#4$\sim$\#8, using the extracted relations, the proposed ASG for all models was successful, as in the case with \#2.

Subsequently, we applied the proposed ASG to models \#1, \#2, and \#3 that have very different parts structures.
\tabref{result_cs3} shows the means plus-minus twice the standard deviations of the maximum fitness values of the generated sequences for the three models.
The percentages of the feasible sequences for all the models are 100\%.
Even when there are multiple part changes in the product, the proposed method can achieve the ASG with a high reproducibility.
%%%%%%%%%%%%%%%%%%%%%%%%%%%%%%%%%%
\begin{table}[tb]
    \centering
        \caption{\small{Reproducibility of ASG determined in Case Study~3. Each element shows the mean plus-minus twice the standard deviation of the calculated fitness values. The percentage of feasible sequences for the evaluated models is 100\%.}} \tablab{result_cs3}
        \begin{tabular}{crrr} \toprule
            \begin{tabular}{c}Model\end{tabular}
            & \begin{tabular}{c}Fitness~1\end{tabular}
            & \begin{tabular}{c}Fitness~2\end{tabular}
            & \begin{tabular}{c}Sum of 1 and 2\end{tabular} \\ \midrule
            \#1 & $9.50\pm{1.00}$ & $28.0\pm{8.00}$ & $37.5\pm{7.00}$ \rule[-1mm]{0mm}{4mm} \\
            \#2 & $51.7\pm{7.32}$ & $356\pm{5.38}$ & $408\pm{5.52}$ \rule[-1mm]{0mm}{4mm} \\
            \#3 & $58.7\pm{6.06}$ & $415\pm{2.40}$ & $473\pm{6.94}$ \rule[-1mm]{0mm}{4mm} \\
            \bottomrule
        \end{tabular}
\end{table}
%%%%%%%%%%%%%%%%%%%%%%%%%%%%%%%%%%
\begin{figure}[tb]
  \begin{minipage}[tb]{\linewidth}
    \centering
    \includegraphics[width=0.7\linewidth]{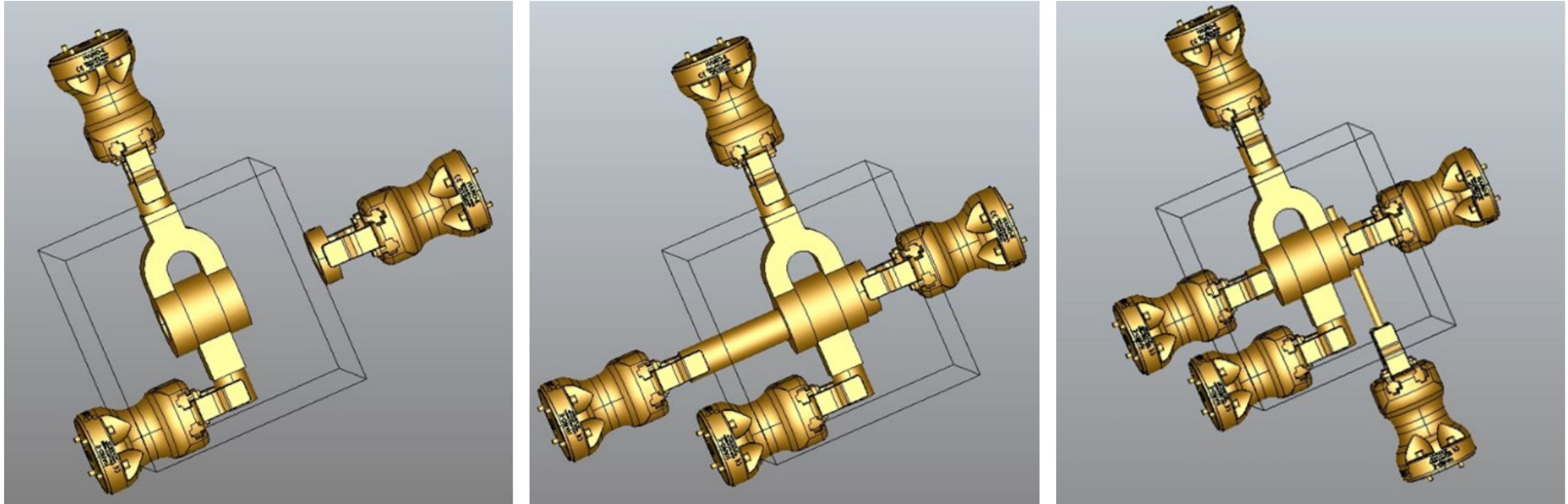}
    \subcaption{\small{Interference-free sequence}}
  \end{minipage}
  \begin{minipage}[tb]{\linewidth}
    \centering
    \includegraphics[width=0.7\linewidth]{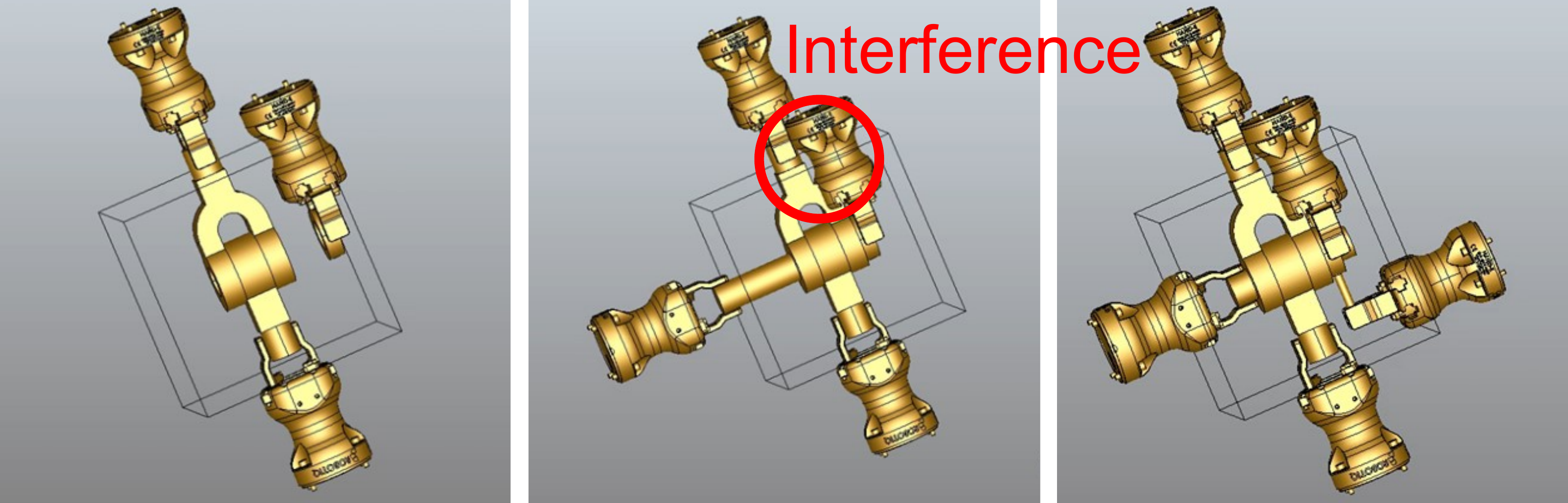}
    \subcaption{\small{Sequence that occurs interference}}
  \end{minipage}
  \caption{\small{Successful (a) and failure (b) simulation examples with a robotic gripper model for graspable sequences.}}
  \figlab{sim-grasp}
\end{figure}
%%%%%%%%%%%%%%%%%%%%%%%%%%%%%%%%%%
\begin{figure}[tb]
  \centering
  \begin{minipage}[tb]{\linewidth}
    \centering
    \includegraphics[width=0.9\linewidth]{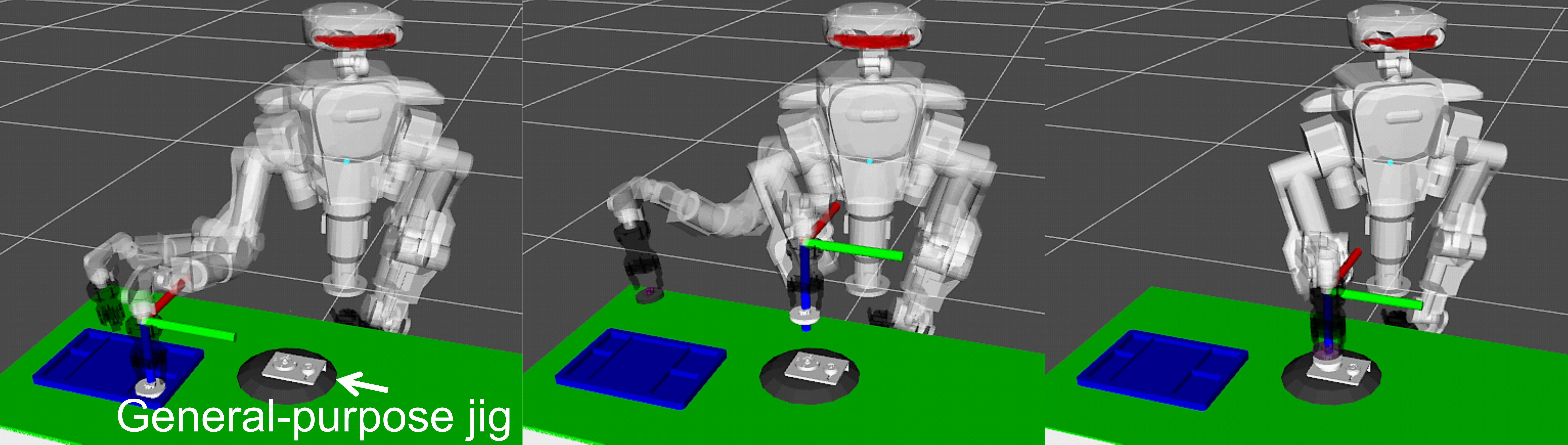}
    \subcaption{\small{Pick-and-place of an idler (P11 shown in~\figref{case2})}}
  \end{minipage}
  \begin{minipage}[tb]{\linewidth}
    \centering
    \includegraphics[width=0.9\linewidth]{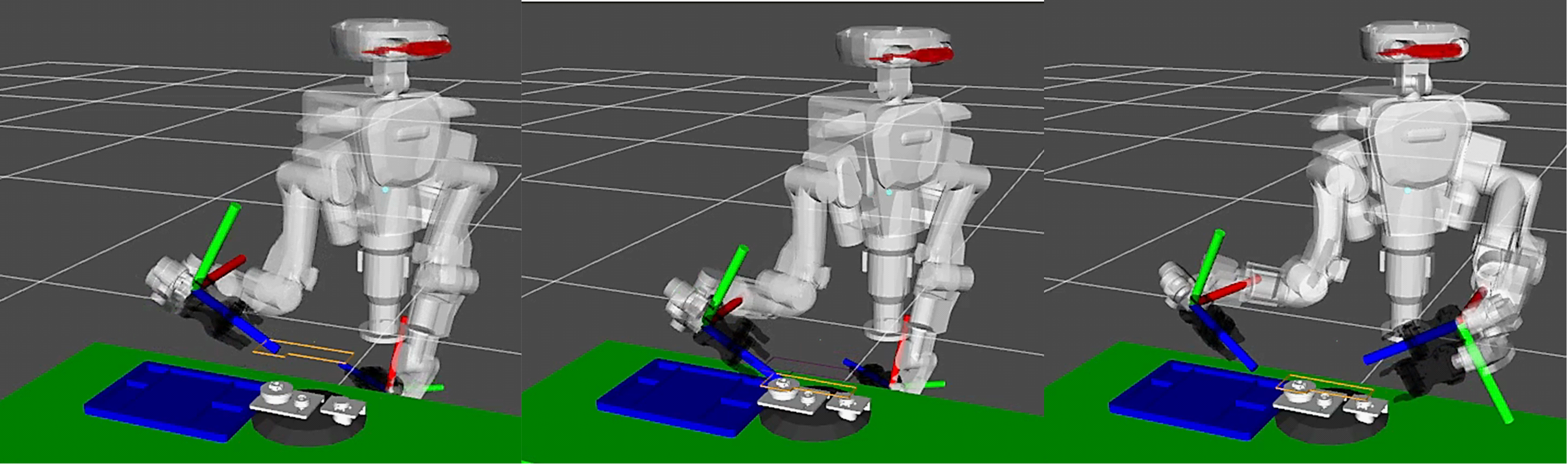}
    \subcaption{\small{Pick-and-place of a rubber band (P6 shown in~\figref{case2})}}
  \end{minipage}
  \caption{\small{A succeessful simulation of robot motions with the graspable sequence.}}
  \figlab{sim-robot}
\end{figure}
%%%%%%%%%%%%%%%%%%%%%%%%%%%%%%%%%%

\section{Discussion}
\subsection{Extensibility on Handling Deformable Parts}
For string-like deformable parts with snap-fit plugs, the assembly direction of the plug would be erroneously determined as interference.
It is thus necessary to be able to recognized a snap-fit connector as an object that can be assembled, based on the CAD geometry~\cite{Shellshear2020}. For ring-shaped deformable parts, an assembled CAD model that was deformed is necessary. 
The extent to which this deformation is represented in the model depends on the product designer. 

Ghandi~\etal~\cite{Ghandi2015} used \textit{Finite Element Method} (FEM) simulation for the ASG for deformable parts.
For the FEM, a user must identify the property of assembly parts. 
The manual measurement of the property is time-consuming and the accuracies influence the results of ASG.
To replace such a method, we will develop a time-efficient ASG method based on the geometries and semantic information of parts.

\subsection{Graspable Sequences Toward Grasp Planning}
Once the assembly sequence is determined, the feasible grasping based on interference between the robot end effector and parts must be determined.
\figref{sim-grasp} shows the process for determinating the grasping points and interference in a sequence generated in Case Study~1. 

The following procedure was used.
\begin{enumerate}
   \item Randomly sampling hand-crafted graspable points on the object surface
   \item Generating concatenated models of the parts and the gripper by fixing a certain pose of the gripper
   \item Determining the interference by moving the concatenated models in the simulation using CAD models
\end{enumerate}
To achieve robotic grasping, such as by using the CAD-based method, we can determine the occurrence of interference. 

\subsection{Application to Industrial Robotic Assembly}
To clarify the limitations of the assembly sequence generated in Case Study~2, we simulated the handcrafted assembly operations shown in~\figref{sim-robot}.

In fact, after the insertion of the rubber band, the robot needs to support the non-fixed parts.
Based on the center of gravity of the parts (\eg~\cite{Murali2019,Gulivindala2020}), we must fix the parts in stable positions at some point by using assembly jigs. 
Since the preparation of custom-made jigs is labor-intensive, we used the \textit{Soft jig}~\cite{Softjig} to fix all the parts.
We could therefore complete the operation using the jig.

Compared to the serial assembly sequence discussed in this study, a parallel assembly sequence divided into sub-assemblies (\eg~\cite{Bahubalendruni2019,Yang2020}) is more time-efficient. 
The parallel one would be applicable for industry use.
For example, the driven side (blue frame) and drive side (red frame) of the serial sequence shown in~\figref{sequence_cs2} can be parallelized.

\section{Conclusion}
To generate easy-to-handle assembly sequences for robots, this study addressed assembly sequence generation by considering two tradeoff objectives: (1) insertion conditions and (2) degrees of constraints among assembled parts. 
We propose a multiobjective genetic algorithm to balance these two objectives. 
Furthermore, by deforming the 3D model, the proposed method enables the extraction of two-part relationships based on the displacement of a deformable object as well as a rigid body. 

The interference-free, insertion, and degree of constraint matrices for deformable parts in eight models were successfully extracted. 
The proposed ASG succeeded in generating the sequence for many parts that include deformable parts. 
The ASG with robot motion planning, as proposed in~\cite{Wan2018,Harada2019,Wan2021}, would be a promising direction for future study.

\bibliographystyle{IEEEtran}
\footnotesize
\bibliography{bibliography/reference}

% that's all folks
\end{document}